\documentclass[preprint,12pt,authoryear]{elsarticle}

\usepackage[utf8]{inputenc}
\usepackage[T1]{fontenc}
\usepackage{amsmath,amssymb,amsfonts}
\usepackage{graphicx}
\usepackage{booktabs}
\usepackage{multirow}
\usepackage{array}
\usepackage{tabularx}
\usepackage{makecell}
\usepackage{caption}
\usepackage{subcaption}
\usepackage{rotating}
\usepackage{stfloats}
\usepackage{float}
\usepackage{placeins}
\usepackage{algorithm2e}
\usepackage[colorlinks=true,linkcolor=black,citecolor=blue,urlcolor=blue,breaklinks=true]{hyperref}
\usepackage{xurl}

\journal{Artificial Intelligence in Agriculture}

\begin{document}

\begin{frontmatter}

\title{The Turning Point of 3D Plant Phenotyping: 3D Foundation Models Enable Minute-to-Second Cross-Crop Reconstruction and Beyond}

\author[aff1]{Hanyue Jia\fnref{equal}}
\author[aff2]{Wei Zhou\fnref{equal}}
\author[aff3]{Wenbo Zhou}
\author[aff3]{Yanan Li}
\author[aff2]{Hao Lu\corref{cor1}}
\ead{hlu@hust.edu.cn}
\author[aff1]{Tingting Wu\corref{cor2}}
\ead{tt\_wu@nwsuaf.edu.cn}

\fntext[equal]{These authors contributed equally to this work.}
\cortext[cor1]{Corresponding author.}
\cortext[cor2]{Corresponding author.}

\affiliation[aff1]{organization={Northwest A\&F University},
    city={Yangling},
    postcode={712100},
    country={China}}

\affiliation[aff2]{organization={Huazhong University of Science and Technology},
    city={Wuhan},
    postcode={430074},
    country={China}}

\affiliation[aff3]{organization={Wuhan Institute of Technology},
    city={Wuhan},
    postcode={430205},
    country={China}}

\begin{abstract}
3D plant phenotyping is notoriously known to be procedure-complicated and of low throughput due to the extensive multi-view imaging, the fragile 3D reconstruction pipeline, and the additional cost from reconstructed geometry to phenotypic extraction. These limitations are further amplified in low-cost data acquisition, where smartphone videos or sparsely sampled multi-view images provide limited view overlap and self-occlusion. In this work, we show that the conventional 3D plant phenotyping pipeline could be streamlined and significantly accelerated with 3D Foundation Models (3DFMs), and particularly, present one of the first cross-crop 3D phenotyping frameworks powered by 3DFMs. The framework replaces COLMAP-style sparse initialization with 3DFM-based feed-forward geometric recovery, combines geometry-constrained 3D Gaussian Splatting for dense reconstruction, enables few-view reconstruction through iterative view synthesis and refinement, and converts reconstructed geometry into measurable organs through 2D-to-3D semantic transfer, metric scale recovery, and organ instance separation. We further construct a cross-crop dataset with smartphone-based image acquisition, diverse plant morphologies, and manual annotations for segmentation and phenotypic evaluation. Experiments across $26$ plant sequences show that 3D Foundation Models reduce the average reconstruction time from $6.52$ minutes to $1.58$ seconds while maintaining high reconstruction quality and phenotyping accuracy. These results suggest a fresh technical route for high-throughput 3D plant phenotyping, from low-cost image acquisition to fast reconstruction, perception, scale recovery, and phenotypic measurement.
\end{abstract}


\end{frontmatter}


\section{Introduction}
Plant phenotype
links between genotype and environment~\citep{houle2010phenomics}.
To acquire reliable plant phenotypes, extensive sensing technologies have been used, particularly image-based sensing~\citep{hu2025openpheno,tao2022proximal}.
Compared with 2D imaging, 3D phenotyping captures canopy architecture, organ geometry, and plant structure more directly,
making it possible to quantify geometric traits such as leaf orientation, light interception, and architectural organization~\citep{okura2022plants3dreview,yang2026threeDphenotyping}. As high-throughput phenotyping continues to expand, a central challenge is how to acquire quantitatively usable 3D plant structure at low cost and high efficiency~\citep{furbank2011phenomics,yang2020cropphenomics}.

Early 3D plant phenotyping relied mainly on active sensing systems, including LiDAR, structured light, and time-of-flight cameras~\citep{okura2022plants3dreview}. These platforms can provide accurate depth measurements and dense point clouds, but their cost, deployment complexity, and downstream processing requirements have limited broad adoption. To lower the hardware barrier, later studies increasingly turned to passive multi-view reconstruction using consumer cameras or smartphones~\citep{li2025plantreconsurvey,zheng2026mobilepheno3d}. Structure from Motion (SfM) and Multi-View Stereo (MVS) offered a practical route toward low-cost 3D modeling~\citep{schonberger2016sfm,furukawa2010mvs}, yet these pipelines rely on local feature extraction, cross-view matching, incremental pose estimation, and dense correspondence search, making them computationally time-consuming and strongly dependent on sufficient image overlap. In plant scenes, such dependence is further challenged by repeated texture, severe self-occlusion, and thin structures~\citep{harandi2023sense3d}. Under sparse-view or rapid-acquisition conditions, classic pipelines often fail during camera initialization or produce unstable geometry, which limits their usefulness for high-throughput phenotyping~\citep{schonberger2016sfm}.

Neural rendering has
recently advanced image-based 3D reconstruction~\citep{mildenhall2020nerf}. Neural Radiance Fields demonstrated the potential of continuous scene representation, and 3D Gaussian Splatting (3DGS) greatly improved rendering efficiency, becoming a major explicit representation framework~\citep{kerbl2023gaussiansplatting}. Subsequent variants, including SuGaR, 2DGS, and Mip-Splatting, improved surface alignment, geometric consistency, and anti-aliasing~\citep{yu2024mipsplatting}.

Plant-oriented studies have also begun to connect neural or point-based 3D representations with phenotypic analysis. PlantSegNeRF~\citep{yang2025plantsegnerf} explores NeRF/3DGS-based instance extraction, IPENS~\citep{song2025ipens} focuses on interactive phenotypic analysis, Wheat3DGS~\citep{zhang2025wheat3dgs} investigates crop-specific organ measurement, PlantGaussian~\citep{shen2025plantgaussian} extends 3D Gaussian splatting to cross-time and cross-scene plant representation, Hyperspectral Imaging Meets 3D Gaussian Splatting~\citep{deng2026hyperspectral3dgs} further pushes plant-oriented 3DGS beyond pure morphology toward multimodal structural--spectral representation, and Eff-3DPSeg~\citep{luo2023eff3dpseg} addresses weakly supervised point-cloud segmentation. These studies show that 3D reconstruction is moving toward plant perception and measurement, but most remain focused on specific crops, organs, or segmentation tasks. More importantly, standard 3DGS still depends on COLMAP-style sparse initialization, so the fragile front end is largely unchanged~\citep{kerbl2023gaussiansplatting}. Its optimization target also remains primarily photometric, whereas plant phenotyping requires geometrically reliable representations of thin leaves, fine stems, and open boundaries for downstream measurement~\citep{ojo2024splanting}.

Recently
3D foundation models (3DFMs) emerge and provide a new entry point for overcoming this front-end

bottleneck~\cite{el2024probing}.
Their progress is not simply a matter of applying larger models to reconstruction, but reflects a shift in how multi-view geometry is inferred. DUSt3R~\citep{wang2024dust3r} moved multi-view recovery from feature matching and triangulation toward feed-forward point-map prediction, and MASt3R~\citep{leroy2024mast3r} further coupled matchability with geometric consistency. VGGT~\citep{wang2025vggt} unified camera parameters, point maps, depth maps, and point tracks within a single model, making it possible to obtain a more complete geometric initialization in one forward pass. $\pi^3$~\citep{wang2025pi3} further removed the fixed-reference-view assumption and improved robustness to input ordering through a reference-free, permutation-equivariant design. Recent methods such as FLARE~\citep{zhang2025flare} and AMB3R~\citep{wang2025amb3r} extend this trend toward joint recovery of cameras, geometry, and appearance under sparse-view settings.
In short, these methods shift front-end
geometric
recovery from local matching-based incremental optimization to rapid geometric inference driven by learned multi-view priors, creating a practical opportunity for second-level camera initialization and initial structure recovery in plant scenes~\citep{xu2022smartbreeding}.

Nevertheless, existing progress has not yet overcome the front-end constraint in low-cost image-based 3D plant phenotyping. Both standard 3DGS and most image-based reconstruction pipelines still depend on COLMAP-style initialization. In plant scenes, its local feature matching and incremental pose estimation are not only time-consuming but frequently become the primary failure point under sparse-view or rapid-acquisition conditions.
In contrast, 3DFMs provide a new opportunity to move beyond this matching-based front end, but their ability to stably replace COLMAP on cross-crop plant data, reduce the number of views required for usable reconstruction, and still support downstream organ-level phenotypic measurement remains insufficiently validated in the plant domain~\citep{wang2025vggt,wang2025pi3}. Meanwhile, existing plant-oriented 3D methods have begun to integrate reconstruction with segmentation and phenotypic analysis, yet they often rely on already available 3D results or focus on specific crops, organs, and processing stages~\citep{yang2025plantsegnerf,song2025ipens,zhang2025wheat3dgs,luo2023eff3dpseg}. The key gap therefore is to
explore whether
3DFMs can provide
measurement-ready geometric initialization under low-cost, few-view acquisition conditions and support a measurement-oriented workflow from dense representation, semantic interpretation, and scale recovery to organ-level trait
extraction.

\begin{figure*}[!t]
    \centering
    \includegraphics[width=\linewidth]{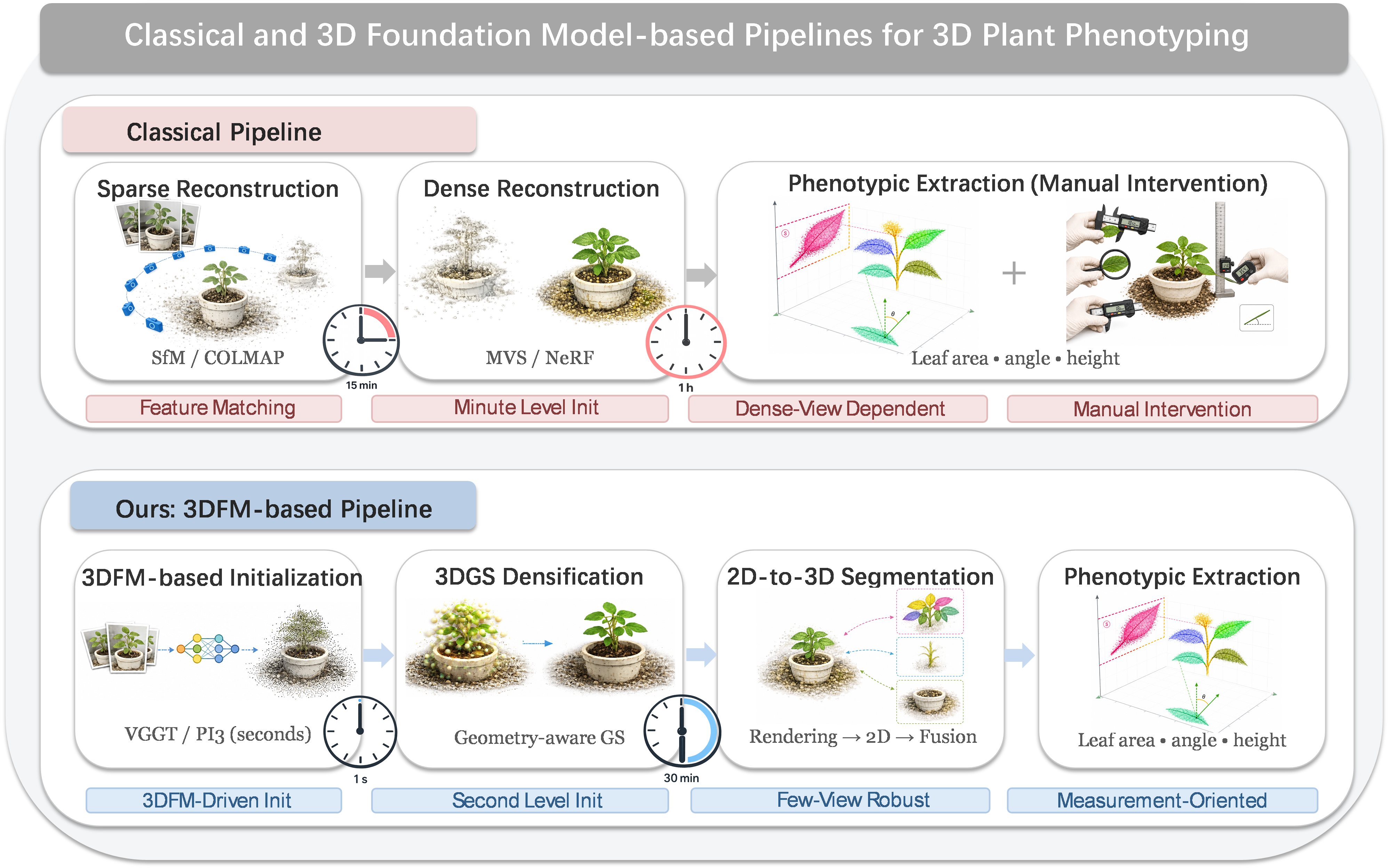}

    \caption{\textbf{Comparison between classical and 3DFM-based pipelines for 3D plant phenotyping.}
    Classical pipelines typically proceed from image acquisition to COLMAP/SfM-based initialization, dense reconstruction, and manual or interactive trait extraction, and are limited by minute-level initialization, dense-view dependence, and manual intervention.
    The proposed pipeline replaces the matching-based front end with 3DFM-based second-level initialization, followed by 3DGS densification, 2D-to-3D segmentation, and organ-level phenotypic extraction.
    This design converts low-cost image inputs into measurable 3D plant traits through a faster and more measurement-oriented reconstruction workflow.}
    \label{fig:pipeline}
\end{figure*}

In this work, we propose to replace COLMAP-style sparse initialization with
3DFMs for 3D plant phenotyping, achieving second-level recovery of camera parameters and initial structure~\citep{wang2025vggt,wang2025pi3}. This initialization is then converted into a dense 3D representation that is better suited for delineating leaf boundaries, fine stems, and canopy structures through geometry-constrained 3DGS~\citep{yu2024mipsplatting}. Under few-view conditions, iterative view synthesis and refinement are employed to supplement effective observations~\citep{long2022sparseneus,liu2023zero123}. Finally, 2D-to-3D semantic transfer, metric scale recovery, and
organ instance separation convert the reconstructed geometry into measurable
organ-level instances. In this way, 3DFMs, 3DGS, 3D perception, and phenotypic computation serve a unified goal:
\textit{the rapid generation of measurable 3D plant structures}.

Experiments across different crop species and varying acquisition conditions show that
3DFMs advance the front end of 3D plant phenotyping from the minute level to the second level while preserving the geometric consistency required for organ-level measurement. Across $26$ plants, the 3DFM-based front-end initialization reduces the average time from $6.52$ minutes to $1.58$ seconds, with comparable reconstruction quality.
Across representative multi-crop samples, the ratio between estimated and measured total leaf area remains within  $0.9514$--$1.0629$, and the mean absolute error of leaf inclination angle is approximately $2.04^\circ$. These results indicate that
3DFMs not only accelerate plant 3D reconstruction, but can also serve as a geometric entry point for low-cost cross-crop phenotyping, supporting a continuous transition from reconstruction to perception and organ-level trait extraction~\citep{okura2022plants3dreview}.

The main contributions of this study
include the following.
\begin{enumerate}

    \item \textbf{We systematically validate 3DFMs as second-level front-end initializers for plant 3D phenotyping.}
    VGGT and $\pi^3$ compress the COLMAP-style front end, including feature matching, pose estimation, and sparse reconstruction, into feed-forward geometric prediction, simplifying and accelerating initialization from minutes to seconds.

    \item \textbf{We propose a 3DFM--3DGS framework that lowers the view threshold for plant reconstruction.}
    The framework combines 3DFM initialization, 3DGS densification, and view supplementation, enabling few-view inputs to support usable reconstruction and phenotypic measurement.

    \item \textbf{We
    	contributed a cross-crop 3D phenotyping dataset with organ-level ground truths.}
    The dataset covers diverse crops, morphologies, environments, and growth stages, and provides leaf-instance annotations and manual trait measurements for joint evaluation of reconstruction, perception, and measurement.
\end{enumerate}

\section{Materials and methods}

\subsection{Data acquisition}

\begin{figure*}[!t]
    \centering
    \includegraphics[width=1.0\linewidth]{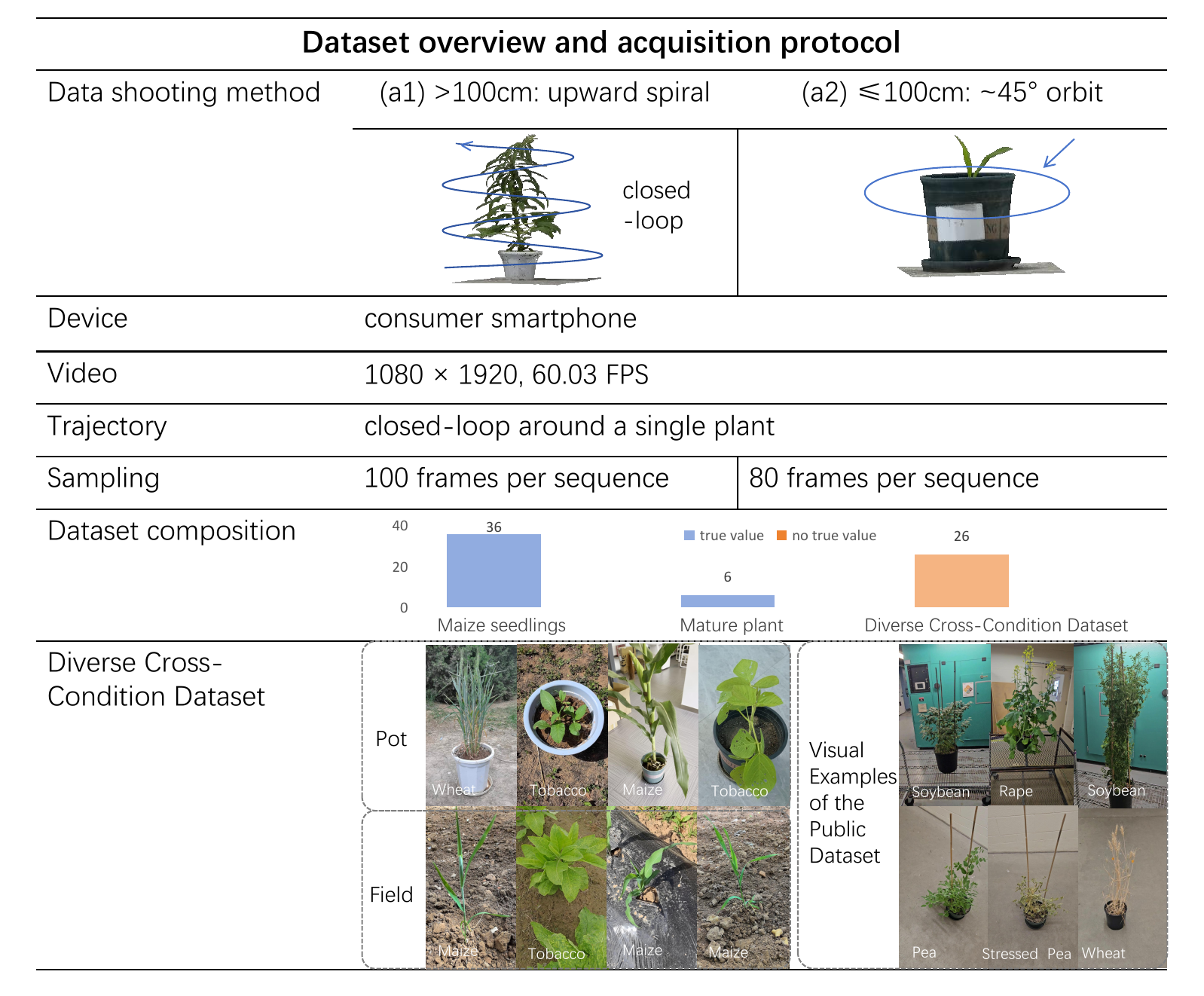}
    \caption{\textbf{Dataset overview and acquisition protocol.} Smartphone videos were acquired along closed-loop trajectories with height-adaptive shooting and frame sampling. The dataset included a robustness subset spanning multiple species and conditions, and a quantitative subset with manual annotations for segmentation and phenotypic evaluation.}
    \label{fig:dataset_overview}
\end{figure*}

To support the evaluation of reconstruction efficiency, few-view robustness, 3D semantic segmentation, and organ-level phenotypic measurement, we constructed a smartphone-based cross-crop dataset combining multi-view video acquisition, fixed-budget frame sampling, and manual ground-truth annotation. All sequences were captured using a consumer smartphone with a resolution of 1080 $\times$ 1920 and a frame rate of 60.03 FPS under variable illumination. Each video was recorded around a single plant along a continuous closed-loop trajectory to ensure sufficient viewpoint coverage for camera pose estimation and 3D reconstruction. To balance reconstruction completeness against computational cost, a height-adaptive sampling strategy was used, with 100 frames retained for plants taller than 100 cm and 80 frames for plants of 100 cm or less. The overall acquisition protocol and dataset composition are summarized in Fig.~\ref{fig:dataset_overview}.

The dataset comprised a robustness subset and a quantitative evaluation subset. The robustness subset contained 26 sequences named by species, acquisition condition, and sequence index, such as Tobacco\_Outdoor\_01, Maize\_Field\_01, and Wheat\_Indoor\_01. This subset was used to assess performance across species, growth stages, acquisition environments, and structural complexity. It covered maize, tobacco, wheat, soybean, bamboo, rapeseed, legume, pea, broccoli, and related crops under indoor, outdoor potted, and field conditions. Among them, eight sequences were derived from the public dataset Splanting: 3D plant capture with Gaussian splatting~\citep{ojo2024splanting}. The quantitative evaluation subset consisted of a mature-plant subset and a temporal maize-seedling subset. The mature-plant subset included five plants from soybean, sesame, and maize, with manual leaf-instance annotations and leaf-level phenotypic ground truth for segmentation and trait evaluation. The temporal subset included nine maize seedlings recorded at four time points, yielding 36 video sequences for stage-wise phenotypic evaluation.

Acquisition trajectories were adjusted according to plant height. Plants taller than 100 cm were recorded using an upward spiral trajectory from the basal stem to the canopy, whereas plants of 100 cm or less were captured using a circular trajectory with an approximately 45$^\circ$ downward viewing angle. In all cases, trajectory continuity and loop closure were maintained to ensure complete plant coverage. For sequences used for absolute phenotypic measurement, the pot diameter was measured before imaging and the pot rim was kept visible to provide an in-scene metric reference for scale recovery.

Manual measurements were collected for plant height, leaf length, leaf width, leaf area, and leaf inclination angle. Leaf length, width, and area were measured using the OpenPheno~\citep{hu2025openpheno} mini-program, whereas leaf inclination angle was measured using a leaf inclination meter.
To ensure consistency with algorithmic outputs, leaf length was defined along the major axis, leaf width as the maximum transverse width, and leaf area as the one-sided area of a single leaf blade. Leaf inclination angle was defined as the acute angle between the fitted mean leaf plane and the horizontal plane. In the subsequent 3D computation, this quantity is equivalently expressed as the acute angle between the mean leaf surface normal and the global vertical direction.
Repeated measurements were averaged, and the corresponding variance was recorded as an estimate of annotation noise~\citep{itakura2019leafinclination}.
All experiments were conducted on a Linux workstation equipped with an NVIDIA RTX A6000 GPU with 48 GB memory. The software environments followed the official configurations of VGGT and \(\pi^3\), based on Python 3.10 or later and CUDA-enabled PyTorch. VGGT, \(\pi^3\), SAM, and Difix3D+ were used with public pretrained weights without further fine-tuning. The 3DGS/Mip-Splatting optimization was performed for 30,000 iterations using default settings unless otherwise specified. Novel-view augmentation was used only in the few-view experiments and ablation studies.

\subsection{3DFM-based rapid initialization}

Front-end initialization recovers camera parameters and initial geometry as the common starting point for densification, semantic transfer, and phenotypic measurement. In plant scenes, it must also provide cross-view consistency in a unified coordinate system. Conventional SfM/COLMAP obtains this through local feature extraction, matching, pose estimation, and sparse reconstruction~\citep{schonberger2016sfm}, but repetitive textures, thin organs, occlusions, and uneven views reduce matching stability and increase cost~\citep{okura2022plants3dreview,harandi2023sense3d}. This stage therefore targets rapid recovery of stable cameras and initial geometry.

We use visual-geometry 3D foundation models (3DFMs) as feed-forward initializers. Unlike SfM, which reconstructs cameras and structure from local correspondences, 3DFMs infer multi-view geometry directly from images with learned priors. Following DUSt3R and MASt3R~\citep{wang2024dust3r,leroy2024mast3r}, we evaluate VGGT and $\pi^3$ as representative initializers~\citep{wang2025vggt,wang2025pi3}. VGGT predicts camera and geometric attributes jointly, whereas $\pi^3$ uses a reference-free, permutation-equivariant formulation. Both share the same conversion interface, and $\pi^3$ is used as the default initializer unless otherwise specified.

Let the sampled multi-view images be denoted as
\[
I=\{I_i\}_{i=1}^{N}, \qquad
O_{\mathrm{raw}}^{m}=F_{\mathrm{3DFM}}^{m}(I), \quad m\in\{\mathrm{VGGT},\pi^3\}.
\]
Here, \(O_{\mathrm{raw}}^{m}\) denotes the model-specific raw geometric outputs, including camera predictions, view-wise geometric maps, depth-related quantities, and confidence or validity information. Because these outputs differ across models, \(O_{\mathrm{raw}}^{m}\) serves as a unified notation rather than assuming identical formats for VGGT and $\pi^3$.

The downstream pipeline requires standardized cameras and an initial sparse point cloud rather than raw point or depth maps. We therefore introduce a bridge conversion step:
\[
(\Theta,P_s)=F_{\mathrm{convert}}(O_{\mathrm{raw}}^{m}),
\qquad
\Theta=\{(K_i,T_i)\}_{i=1}^{N}.
\]

This conversion reorganizes camera predictions into a unified intrinsic--extrinsic representation, maps model-specific geometry into a common coordinate frame, filters unreliable predictions, and fuses the remaining geometry into \(P_s\). The resulting \((\Theta,P_s)\) is comparable to COLMAP initialization for 3DGS, but is obtained by feed-forward 3DFM inference rather than matching-based incremental reconstruction. It provides reconstruction-scale initialization, while absolute metric scale is recovered later using the in-scene pot reference.

Through this interface, VGGT and $\pi^3$ serve as interchangeable 3DFM front ends. The standardized \((\Theta,P_s)\) provides direct geometric input for Mip-Splatting-based densification, 2D-to-3D semantic transfer, and organ-level phenotypic measurement.

\subsection{Geometry-constrained 3DGS densification}

The 3DFM initialization result \((\Theta,P_s)\) provides stable camera parameters and initial sparse geometry, but it is still insufficient for the semantic and geometric operations required by organ-level measurement. For plant scenes, the 3D representation must cover organ structures more completely and preserve geometric continuity around leaf boundaries, thin stems, and overlapping regions. This stage therefore transforms the standardized initialization into a dense representation. The optimized Gaussian representation \(G\) supports differentiable rendering, whereas the extracted dense point cloud \(P_d\) serves as the discrete geometric carrier for 2D-to-3D semantic transfer, scale recovery, leaf instance separation, and local trait computation.

3DGS provides an efficient explicit representation for converting sparse initialization into a continuous 3D scene representation~\citep{kerbl2023gaussiansplatting}. However, standard photometric 3DGS may produce aliasing, boundary dilation, and floating artifacts around thin plant structures. Since plant phenotyping requires both smooth leaf surfaces and well-preserved high-frequency structures such as leaf edges, leaf tips, and thin stems, we adopt Mip-Splatting as the geometry-constrained 3DGS densification method ~\citep{yu2024mipsplatting,ojo2024splanting} in the main pipeline. Its 3D smoothing and 2D footprint control improve geometric stability under scale variation while maintaining efficient rendering.

Given the camera parameter set \(\Theta=\{(K_i,T_i)\}_{i=1}^{N}\) and the initial sparse point cloud \(P_s=\{p_j\}_{j=1}^{M}\), Gaussian primitives \(g_j=(\mu_j,\Sigma_j,\alpha_j,c_j)\) are initialized from \(P_s\), where \(\mu_j\), \(\Sigma_j\), \(\alpha_j\), and \(c_j\) denote the center, covariance, opacity, and color of the \(j\)-th Gaussian. The representation is then optimized under multi-view photometric constraints:
\[
G^{(0)}=\operatorname{Init}(P_s), \qquad
G=\operatorname*{arg\,min}_{G}\sum_{i=1}^{N}
\mathcal{L}_{\mathrm{photo}}\!\left(R(G;K_i,T_i),I_i\right),
\]
where \(R(\cdot)\) denotes differentiable rendering and \(\mathcal{L}_{\mathrm{photo}}\) denotes the photometric reconstruction loss.

To stabilize Gaussian scales and image-space projections, Mip-Splatting constrains both the 3D covariance and the projected 2D footprint:
\[
\widetilde{\Sigma}_j=\Sigma_j+\sigma_{3D}^{2}I_3, \qquad
\widetilde{\Sigma}^{2D}_{ij}=J_{ij}\widetilde{\Sigma}_jJ_{ij}^{\top}+\sigma_{2D}^{2}I_2,
\]
where \(\sigma_{3D}\) and \(\sigma_{2D}\) control the minimum smoothing scale in 3D space and image space, respectively, and \(J_{ij}\) is the projection Jacobian of Gaussian \(g_j\) in view \(i\). The 3D term suppresses degenerate or excessively sharp primitives, while the 2D term reduces aliasing and boundary dilation in projection.

During optimization, Gaussians with large image-space gradients are densified to improve local coverage around undersampled leaf edges, leaf tips, and thin stems, whereas low-opacity, unstable, or isolated primitives are pruned to suppress floating artifacts. After optimization, a dense point cloud is extracted from the Gaussian representation:
\[
P_d=S(G).
\]
In practice, \(S(\cdot)\) extracts valid Gaussian centers after opacity-based filtering and removal of isolated primitives, producing a discrete point cloud with higher spatial coverage and more continuous boundary structures than the original sparse initialization.

The resulting \(G\) and \(P_d\) provide complementary outputs for the subsequent pipeline: \(G\) supplies the rendering function needed for view-based operations, while \(P_d\) provides the dense geometric support for semantic back-projection, scale recovery, leaf separation, and phenotypic measurement. This 3DGS-centered reconstruction-to-measurement workflow is summarized in Fig.~\ref{fig:integrated_pipeline}.

\subsection{Few-view reconstruction via iterative view synthesis and enhancement}

Few-view reconstruction remains a major bottleneck for high-throughput plant phenotyping~\citep{long2022sparseneus}. Field, handheld, or robotic acquisition may provide only limited views and overlap, causing SfM initialization failures when feature matching becomes unreliable~\citep{schonberger2016sfm}. Neural rendering can also degrade under sparse supervision, producing blurred textures, fragmented leaf geometry, and floating artifacts~\citep{niemeyer2022regnerf}. In plant scenes, thin leaves, repeated texture, self-occlusion, and open organ boundaries further amplify these errors, which may propagate to semantic transfer, scale recovery, leaf instance separation, and phenotypic measurement.

In this study, few-view mainly refers to highly sparse reconstruction with no more than 10 input views, while larger view budgets were evaluated by uniformly subsampling the full closed-loop sequence to characterize the transition toward dense-view performance. The experiment therefore tests both usability under extremely limited input and view-threshold behavior as observations increase. This module was used only for the few-view reconstruction experiments and ablation settings with novel-view augmentation; full-view reconstruction used the original sampled frames without iterative supplementation unless otherwise specified.

Existing few-view methods generally rely on geometric regularization or generative priors. Regularization-based approaches such as SparseNeuS and FreeNeRF stabilize sparse-view optimization~\citep{long2022sparseneus,yang2023freenerf}, but may oversmooth leaf edges, tips, and fine stems. Generative priors can improve visual quality but may introduce cross-view or geometric inconsistencies. Therefore, generated views are used only as additional image observations that provide supplementary photometric constraints, not as geometric ground truth.

\begin{figure*}[!p]
    \centering
    \includegraphics[
        width=\textwidth,
        height=0.88\textheight,
        keepaspectratio
    ]{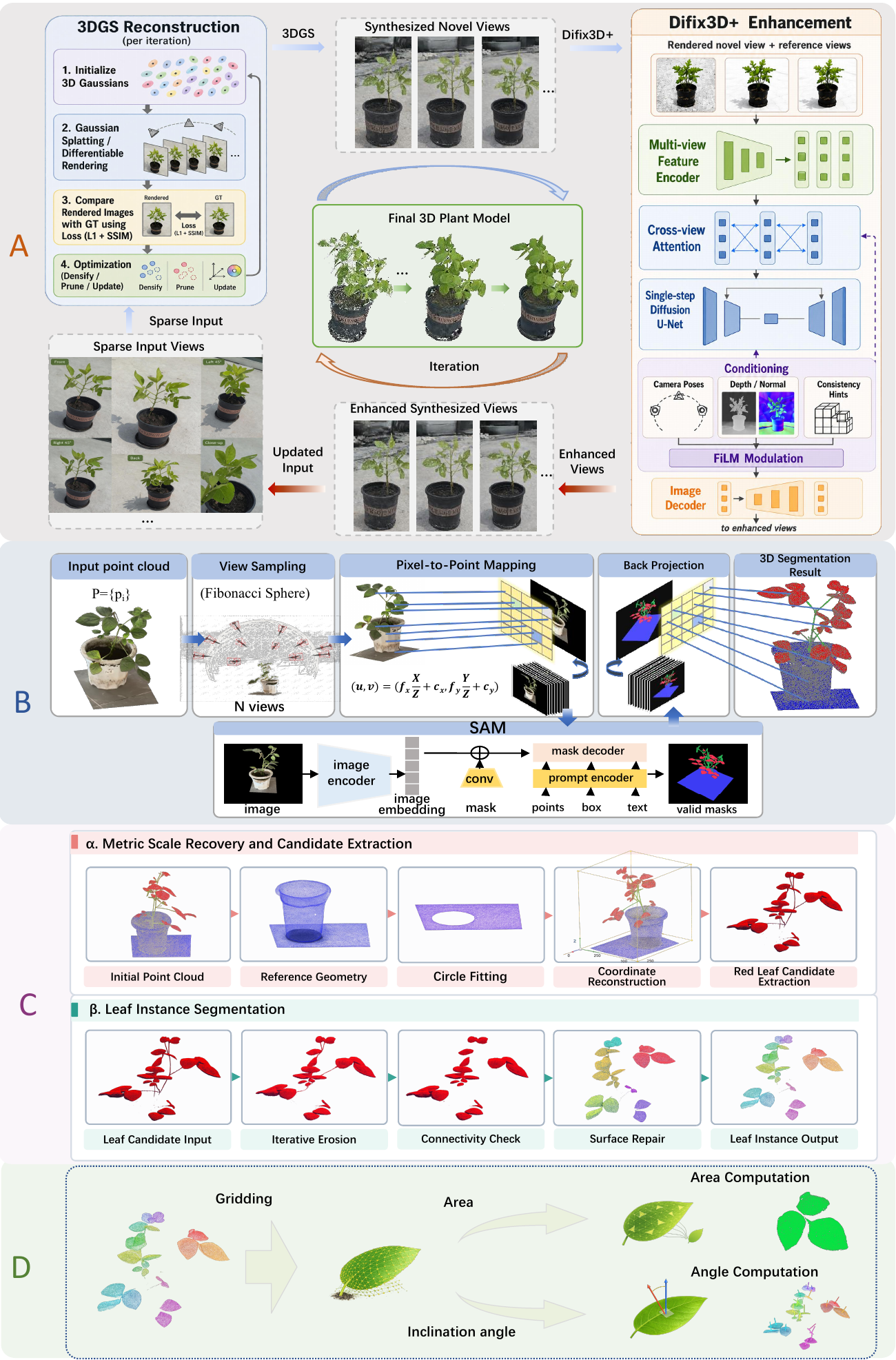}
    \caption{\textbf{Pipeline for few-view 3D plant reconstruction, segmentation, and phenotyping.}
    The workflow starts from sparse multi-view inputs, performs 3DFM-based initialization and Mip-Splatting-based densification, supplements effective observations through Difix3D+-assisted view refinement, and then converts the reconstructed geometry into organ-level traits through 2D-to-3D semantic transfer, scale recovery, leaf instance separation, and phenotypic measurement.}
    \label{fig:integrated_pipeline}
\end{figure*}

Built upon the 3DFM initialization and Mip-Splatting-based densification developed in the previous stages, we introduce an iterative scheme of novel-view synthesis, view refinement, input update, and re-reconstruction, as illustrated in \hyperref[fig:integrated_pipeline]{Fig.~\ref*{fig:integrated_pipeline}a}. Novel-view poses are sampled between existing viewpoints along the closed-loop trajectory, prioritizing large angular gaps. The current Gaussian representation renders intermediate views at these poses, and Difix3D+~\citep{wu2025difix3dplus} refines the rendered images. Difix3D+ is used in its public pretrained form, without further training or fine-tuning, to reduce residual rendering artifacts while using the original sparse views as structural references for geometric consistency.

Let \(I^{(t)}\) denote the input view set at iteration \(t\), \(G^{(t)}\) the current Gaussian representation, and \(\widetilde{\Theta}^{(t)}\) the sampled novel-view poses. The rendering and refinement process is written as
\[
\widetilde{I}^{(t)}=R\!\left(G^{(t)},\widetilde{\Theta}^{(t)}\right),
\qquad
\widehat{I}^{(t)}=F_{\mathrm{Difix}}\!\left(\widetilde{I}^{(t)};I^{(t)}\right),
\]
where \(R(\cdot)\) denotes Gaussian rendering and \(I^{(t)}\) provides the current input views as structural references. The refined views are then merged into the input set and fed back to initialization and densification:
\[
\begin{gathered}
I^{(t+1)} = I^{(t)} \cup \widehat{I}^{(t)},\\
(\Theta^{(t+1)},P_s^{(t+1)}) =
F_{\mathrm{init}}\!\left(I^{(t+1)}\right), \\
G^{(t+1)} =
F_{\mathrm{dens}}\!\left(\Theta^{(t+1)},P_s^{(t+1)}\right).
\end{gathered}
\]
Here, \(F_{\mathrm{init}}\) denotes 3DFM inference followed by the bridge conversion described above, and \(F_{\mathrm{dens}}\) denotes the Mip-Splatting-based densification stage. Since the 3DFM front end accepts variable numbers of input views, the refined novel views can be incorporated into the next reconstruction round.

In practice, novel-view number and location are determined by input sparsity and angular gaps along the acquisition loop, with priority given to weakly observed regions. Iterative rendering, refinement, and re-reconstruction expand sparse inputs into a more informative observation set, improving local structural continuity and organ-boundary completeness before subsequent 2D-to-3D semantic transfer and phenotypic measurement.

\subsection{2D-to-3D semantic transfer}

Direct semantic segmentation on plant point clouds remains challenging due to irregular geometry, thin structures, severe self-occlusion, and the limited availability of large-scale annotated 3D datasets~\citep{schunck2021pheno4d}. To address these issues, we adopt a projection-based 2D-to-3D semantic transfer strategy, which transfers semantic information from image space to 3D space through explicit pixel-to-point correspondence and multi-view label fusion~\citep{imabuchi2024backprojection}, as shown in \hyperref[fig:integrated_pipeline]{Fig.~\ref*{fig:integrated_pipeline}b}.

Given the dense point cloud \(P_d=\{p_i\}_{i=1}^{M}\), where \(p_i\in\mathbb{R}^{3}\), its geometric center and spatial scale are computed as
\[
\bar{p}=\frac{1}{M}\sum_{i=1}^{M}p_i, \qquad
r=\max_i \|p_i-\bar{p}\|.
\]
Based on this normalized geometry, a fixed set of virtual cameras is placed on a sphere with radius proportional to \(r\). Camera positions are generated by Fibonacci sampling to obtain approximately uniform angular coverage. To reduce interference from non-target structures such as pots, views with strong pot dominance or poor plant visibility are avoided, and the remaining viewpoints are biased toward the upper hemisphere of the plant canopy.

For each virtual viewpoint, the point cloud is transformed into the camera coordinate system and projected onto the image plane using a perspective projection model:
\[
u=f_x\frac{X}{Z}+c_x, \qquad
v=f_y\frac{Y}{Z}+c_y.
\]
A depth-based visibility constraint is applied so that only the closest visible point contributes to each pixel. To improve the continuity of rendered images, especially around thin organs, a local splatting strategy is used, allowing each point to influence a small pixel neighborhood. When multiple points contribute to the same pixel, the closest visible point is recorded in the pixel-to-point index map. This index map explicitly stores the correspondence between rendered pixels and 3D points, enabling direct label back-projection without relying on post-hoc nearest-neighbor search~\citep{imabuchi2024backprojection}.

Semantic segmentation is then performed on the rendered images using a SAM-based 2D segmentation module~\citep{kirillov2023sam}.Compared with directly operating on sparse and irregular point clouds, mature 2D segmentation models provide stronger semantic priors for separating plant organs~\citep{chen2018deeplabv3plus,zhang2026depthcropsegpp}. The resulting 2D labels are transferred back to \(P_d\) through the pixel-to-point index map, assigning semantic predictions to the corresponding 3D points.

Since each 3D point may be observed from multiple virtual viewpoints, it can receive multiple semantic predictions. These predictions are aggregated across views, and the final 3D label is determined by majority voting:
\[
L_i=\operatorname*{arg\,max}_{\ell} \sum_k \mathbf{1}\!\left(L_i^{(k)}=\ell\right),
\]
where \(L_i^{(k)}\) denotes the label assigned to point \(p_i\) from the \(k\)-th rendered view, and \(\ell\) denotes the semantic class. This multi-view fusion integrates complementary observations from different directions and reduces the influence of occlusion, local rendering noise, and single-view segmentation errors~\citep{dai2018threedmv}.

The output of this stage is a labeled dense point cloud \((P_d,L_{3D})\), where \(L_{3D}=\{L_i\}_{i=1}^{M}\). This semantic point cloud provides the direct input for subsequent scale recovery, leaf instance separation, and organ-level phenotypic measurement.

\subsection{Scale recovery and leaf instance separation}

The labeled dense point cloud \((P_d,L_{3D})\) obtained from 2D-to-3D semantic transfer already carries semantic labels for each point, but it still cannot be used directly for absolute phenotypic measurement. Two issues remain: the reconstruction-scale point cloud lacks physical metric scale, and the semantic leaf class is still a single point set rather than a set of measurable leaf instances. This section therefore performs metric scale recovery for sequences with a measured in-scene reference and separates semantic leaf points into individual leaf instances, corresponding to \hyperref[fig:integrated_pipeline]{Fig.~\ref*{fig:integrated_pipeline}c}.

For sequences used for absolute trait evaluation, the known pot geometry is used as an in-scene metric reference. Conventional scale recovery often relies on external calibration objects, such as checkerboards or reference spheres~\citep{harandi2023sense3d}. These markers increase acquisition complexity and are also prone to occlusion by the plant canopy. In contrast, the pot diameter was measured before imaging and used for scale recovery when the pot rim was reliably visible. Sequences without a reliable metric reference were used for reconstruction-scale analysis rather than absolute physical trait measurement.

We first fit the ground plane with RANSAC and align the vertical direction to the \(Z\) axis, so that the fitted ground plane defines the \(XOY\) plane. The pot-bottom center is then taken as the coordinate origin. The pot region is isolated from the lower part of the reconstructed scene using geometric constraints, and its outer-boundary candidate points are extracted by angular binning in polar coordinates. Circle fitting is then applied to the outer rim of the pot. Let \(D_{\mathrm{real}}\) denote the measured pot diameter and \(D_{\mathrm{rec}}\) the reconstructed diameter estimated from the pot rim. The scale factor and scaled point cloud are defined as
\[
D_{\mathrm{rec}}=F_{\mathrm{pot}}(P_d), \qquad
\hat{s}=\frac{D_{\mathrm{real}}}{D_{\mathrm{rec}}}, \qquad
\widetilde{P}_d=\{\hat{s}(p_i-o)\mid p_i\in P_d\},
\]
where \(o\) denotes the pot-bottom center in the reconstructed coordinate system. After this transformation, \(\widetilde{P}_d\) is represented in a unified physical coordinate system with the \(Z\) axis as the vertical direction, which provides the reference for plant height and leaf inclination measurement.

Based on the scaled dense point cloud and the 3D semantic labels, the points classified as leaf are extracted as
\[
P_{\mathrm{cand}}=\{\tilde{p}_i\in \widetilde{P}_d \mid L_i=\mathrm{leaf}\},
\]
where \(L_i\) denotes the semantic label of point \(p_i\). Because leaves are often connected through petioles, stems, or local contact regions, direct Euclidean clustering or region growing tends to produce severe under-segmentation~\citep{zarei2024plantsegnet}. We therefore adopt a geometry-driven hierarchical separation strategy~\citep{ma2023silique} and write leaf instance separation as
\[
P_{\mathrm{leaf}}=\{P_{\mathrm{leaf}}^q\}_{q=1}^{Q}
=F_{\mathrm{recover}}\!\bigl(F_{\mathrm{conn}}(F_{\mathrm{erode}}(P_{\mathrm{cand}}))\bigr).
\]
Here, \(F_{\mathrm{erode}}\) denotes voxel-based iterative erosion, which removes sparse outliers and weak connecting structures caused by petioles, stems, or local leaf contact; \(F_{\mathrm{conn}}\) denotes connectivity analysis, which partitions the eroded points into leaf-core connected components and cuts residual low-density bridges using local density cues; and \(F_{\mathrm{recover}}\) denotes boundary recovery and point reassignment, which assigns previously removed edge and boundary points back to the nearest compatible leaf core.

Through these steps, the labeled dense point cloud is converted into a scaled plant point cloud and a set of metric leaf instances. The scaled point cloud provides the basis for plant-height estimation, whereas the separated leaf instances provide the direct input for leaf area and leaf inclination angle measurement in the subsequent phenotypic computation stage.

\subsection{Phenotypic measurement}

After scale recovery and leaf instance separation, phenotypic traits are extracted from the scaled plant point cloud and separated leaf instances under a unified physical coordinate system. Let \(P_{\mathrm{leaf}}=\{P_{\mathrm{leaf}}^q\}_{q=1}^{Q}\) denote the separated leaf instances, where \(P_{\mathrm{leaf}}^q\) is the point cloud of the \(q\)-th leaf. For each instance, we reconstruct a 3D mesh and compute leaf area and leaf inclination angle from surface geometry. Plant height is obtained from the scaled dense point cloud as the vertical distance between the pot-bottom reference plane and the highest plant point, and is used as an auxiliary whole-plant trait. The main quantitative evaluation focuses on leaf area and leaf inclination angle, because they directly reflect organ-level geometric completeness and posture recovery; this final trait-extraction stage is shown in \hyperref[fig:integrated_pipeline]{Fig.~\ref*{fig:integrated_pipeline}d}.

For leaf area estimation, we reconstruct an open mesh for each leaf using the Ball-Pivoting Algorithm (BPA)~\citep{bernardini1999bpa}. Compared with closed-surface methods such as Poisson surface reconstruction, BPA is more suitable for thin, open leaf surfaces and reduces overestimation caused by artificial boundary closure. Because leaf phenotyping concerns the one-sided area of a leaf blade, area is computed from the cleaned single-surface mesh after removing duplicated or opposite-side artifacts. Let \(T_q\) denote the triangles in the reconstructed mesh of the \(q\)-th leaf, where \(S_i\) and \(\vec{n}_i=(n_{i,x},n_{i,y},n_{i,z})\) represent the area and normal vector of the \(i\)-th triangle, respectively. Isolated triangles, locally duplicated triangles, and triangles with abnormal area or inconsistent normals are removed. The remaining valid triangle set is denoted as \(\tilde{T}_q\subseteq T_q\). The one-sided area of the \(q\)-th leaf is then defined as
\[
A_q=\sum_{i\in \tilde{T}_q} S_i .
\]

Leaf inclination angle is defined consistently with the manual measurement as the acute angle between the fitted mean leaf plane and the horizontal plane, equivalently computed as the acute angle between the mean leaf surface normal and the global vertical direction~\citep{itakura2019leafinclination}. To reduce local noise and mesh irregularity, triangle normals are oriented consistently within each leaf mesh and then averaged with area weights:
\[
\vec n_q=\frac{\sum_{i\in \tilde{T}_q} S_i\,\vec n_i}{\left\|\sum_{i\in \tilde{T}_q} S_i\,\vec n_i\right\|}.
\]
Let \(\vec v=(0,0,1)\) denote the global vertical vector established during scale recovery. The leaf inclination angle of the \(q\)-th leaf is then computed as
\[
\theta_q=\arccos\left(\left|\vec n_q\cdot \vec v\right|\right).
\]
The absolute value removes normal-direction ambiguity, ensuring a consistent physical reference across plants and leaf instances.

The outputs are the auxiliary whole-plant height and paired leaf-level traits \(\{A_q,\theta_q\}_{q=1}^{Q}\), which are compared with manual measurements in the quantitative evaluation.

\section{Results}

We systematically evaluate the key components of the proposed framework, including fast reconstruction, few-view recovery, segmentation strategy, and phenotypic parameter extraction. Using experiments on multiple crops and diverse scenes, we assess the method's overall performance with respect to reconstruction efficiency, geometric quality, organ-level segmentation, and quantitative measurement accuracy.

\subsection{Overall framework}

To evaluate visual geometry 3D Foundation Models as the front-end geometric foundation for plant 3D phenotyping, this study extends the evaluation scope from reconstruction quality alone to the full phenotyping workflow. The core question is whether low-cost image inputs can achieve second-level stable initialization, and then successfully go through dense reconstruction, semantic perception, scale recovery, and leaf instance separation, ultimately producing comparable organ-level phenotypic traits. The Results section follows the data flow of front-end geometric recovery, few-view reconstruction, 3D semantic perception, terminal phenotypic measurement, providing a staged evaluation of the proposed pipeline.

As illustrated in Fig.~\ref{fig:pipeline}, the framework takes smartphone video frames or sampled multi-view images as input, denoted as \(I=\{I_i\}_{i=1}^{N}\). In the first stage, the 3DFM-based front end, together with bridge conversion, recovers standardized camera parameters and initial sparse geometry:
\[
\begin{aligned}
(\Theta,P_s)&=F_{\mathrm{init}}(I)
=F_{\mathrm{convert}}\!\left(F_{\mathrm{3DFM}}^{m}(I)\right),\\
\Theta&=\{(K_i,T_i)\}_{i=1}^{N}.
\end{aligned}
\]
Here, \(\Theta\) denotes the set of camera intrinsics and poses, \(P_s\) denotes the initial sparse point cloud, and \(F_{\mathrm{init}}\) represents 3DFM inference followed by bridge conversion.

In the second stage, geometry-constrained 3DGS converts the sparse initialization into a continuous 3D representation:
\[
\begin{gathered}
G=F_{\mathrm{dens}}(\Theta,P_s),\\
P_d=S(G).
\end{gathered}
\]
Here, $G$ denotes the Gaussian representation, $P_d$ denotes the dense point cloud extracted from the optimized representation. This stage provides novel-view rendering capacity, more complete leaf boundaries, fine stems, canopy structures, serving as the geometric basis for downstream segmentation, measurement. Under sparse-view input, the framework supplements effective observations through view synthesis, image enhancement, re-reconstruction, reducing the dependence of plant 3D reconstruction on dense acquisition.

In the third stage, the dense 3D result enters the semantic perception process. A 2D-to-3D semantic transfer strategy projects multi-view 2D segmentation results back to 3D space through explicit pixel--point correspondence, then obtains stable 3D semantic labels through multi-view fusion:
\[
L_{3D}=F_{\mathrm{seg}}(P_d;\mathcal{V}).
\]
Here, \(\mathcal{V}\) denotes the virtual viewpoints used for projection, 2D segmentation, back-projection, and multi-view fusion.

After semantic point clouds are obtained, pot geometry is used for metric scale recovery, followed by leaf instance separation to split the leaf class into individual leaf objects:
\[
\begin{gathered}
(\hat{s},\widetilde{P}_d)=F_{\mathrm{scale}}(P_d,D_{\mathrm{real}}),\\
P_{\mathrm{leaf}}=\{P_{\mathrm{leaf}}^q\}_{q=1}^{Q}
=F_{\mathrm{inst}}(\widetilde{P}_d,L_{3D}).
\end{gathered}
\]
The separated metric leaf instances are then used to compute phenotypic traits, including leaf area, leaf inclination angle, and plant height:
\[
\tau
=
F_{\mathrm{trait}}(\widetilde{P}_d,P_{\mathrm{leaf}})
=
\left\{H,\{(A_q,\theta_q)\}_{q=1}^{Q}\right\}.
\]
Through this design, the pipeline of reconstruction, segmentation and measurement forms a continuous data-transformation chain aimed at rapid generation of measurable 3D plant structures from low-cost images.

Based on this framework, the following results address four questions. First, whether 3DFM-based initialization can reduce the conventional minute-level front end to the second level across multi-crop, multi-scene data, while achieving reconstruction quality comparable to COLMAP. Second, whether iterative view synthesis and enhancement can lower the view threshold required for usable reconstruction under limited input views. Third, whether 2D-to-3D semantic transfer can provide stable organ perception on complex plant point clouds. Fourth, whether reconstructed structures, after scale recovery, leaf instance separation, can support quantitative phenotypic measurements such as leaf area, leaf inclination angle. This validation sequence follows the workflow in Fig.~\ref{fig:pipeline}, evaluating the proposed plant 3D phenotyping route in terms of front-end efficiency, geometric stability, semantic perception capacity, and terminal phenotypic accuracy.

\subsection{3DFM-based initialization reduces the plant reconstruction front end from minutes to seconds}

\begin{table*}[!t]
    \centering
    \fontsize{6.8pt}{7.6pt}\selectfont
    \renewcommand{\arraystretch}{1.30}
    \setlength{\tabcolsep}{4.6pt}
    \caption{\textbf{Comparison of reconstruction performance using VGGT, COLMAP, and $\pi^3$ on different plant datasets.} Time is reported in minutes (m) or seconds (s).}
    \label{tab:plant_26_compare}

    \makebox[\textwidth][c]{
    \resizebox{\textwidth}{!}{
    \begin{tabular}{l*{16}{c}}
        \toprule
        \multirow{2}{*}{\raisebox{-0.65ex}{\textbf{Method}}}
        & \multicolumn{4}{c}{\textbf{Tobacco\_Outdoor\_01}}
        & \multicolumn{4}{c}{\textbf{Maize\_Outdoor\_01}}
        & \multicolumn{4}{c}{\textbf{Maize\_Field\_01}}
        & \multicolumn{4}{c}{\textbf{Tobacco\_Outdoor\_02}} \\
        \cmidrule(lr){2-5}
        \cmidrule(lr){6-9}
        \cmidrule(lr){10-13}
        \cmidrule(lr){14-17}
        & {\fontsize{7.2pt}{8.0pt}\selectfont \textbf{SSIM}$\uparrow$}
        & {\fontsize{7.2pt}{8.0pt}\selectfont \textbf{PSNR}$\uparrow$}
        & {\fontsize{7.2pt}{8.0pt}\selectfont \textbf{LPIPS}$\downarrow$}
        & {\fontsize{7.2pt}{8.0pt}\selectfont \textbf{Time}$\downarrow$}
        & {\fontsize{7.2pt}{8.0pt}\selectfont \textbf{SSIM}$\uparrow$}
        & {\fontsize{7.2pt}{8.0pt}\selectfont \textbf{PSNR}$\uparrow$}
        & {\fontsize{7.2pt}{8.0pt}\selectfont \textbf{LPIPS}$\downarrow$}
        & {\fontsize{7.2pt}{8.0pt}\selectfont \textbf{Time}$\downarrow$}
        & {\fontsize{7.2pt}{8.0pt}\selectfont \textbf{SSIM}$\uparrow$}
        & {\fontsize{7.2pt}{8.0pt}\selectfont \textbf{PSNR}$\uparrow$}
        & {\fontsize{7.2pt}{8.0pt}\selectfont \textbf{LPIPS}$\downarrow$}
        & {\fontsize{7.2pt}{8.0pt}\selectfont \textbf{Time}$\downarrow$}
        & {\fontsize{7.2pt}{8.0pt}\selectfont \textbf{SSIM}$\uparrow$}
        & {\fontsize{7.2pt}{8.0pt}\selectfont \textbf{PSNR}$\uparrow$}
        & {\fontsize{7.2pt}{8.0pt}\selectfont \textbf{LPIPS}$\downarrow$}
        & {\fontsize{7.2pt}{8.0pt}\selectfont \textbf{Time}$\downarrow$} \\
        \midrule
        COLMAP
        & 0.893 & 30.372 & 0.130 & 6.189\,m
        & 0.869 & 29.491 & 0.161 & 5.141\,m
        & 0.877 & 28.874 & 0.134 & 8.131\,m
        & 0.901 & 29.784 & 0.128 & 6.862\,m \\
        VGGT
        & 0.875 & 29.347 & 0.165 & 1.865\,s
        & 0.833 & 28.989 & 0.189 & 1.619\,s
        & 0.855 & 27.235 & 0.179 & 1.354\,s
        & 0.865 & 28.913 & 0.167 & 1.427\,s \\
        $\pi^3$
        & 0.889 & 30.218 & 0.133 & 1.602\,s
        & 0.870 & 29.487 & 0.167 & 1.881\,s
        & 0.869 & 28.128 & 0.141 & 0.987\,s
        & 0.891 & 29.297 & 0.134 & 1.390\,s \\
        \bottomrule
    \end{tabular}
    }}

    \vspace{0.42em}

    \makebox[\textwidth][c]{
    \resizebox{\textwidth}{!}{
    \begin{tabular}{l*{16}{c}}
        \toprule
        \multirow{2}{*}{\raisebox{-0.65ex}{\textbf{Method}}}
        & \multicolumn{4}{c}{\textbf{Wheat\_Outdoor\_01}}
        & \multicolumn{4}{c}{\textbf{Maize\_Indoor\_01}}
        & \multicolumn{4}{c}{\textbf{Tobacco\_Field\_01}}
        & \multicolumn{4}{c}{\textbf{Bamboo\_Indoor\_01}} \\
        \cmidrule(lr){2-5}
        \cmidrule(lr){6-9}
        \cmidrule(lr){10-13}
        \cmidrule(lr){14-17}
        & {\fontsize{7.2pt}{8.0pt}\selectfont \textbf{SSIM}$\uparrow$}
        & {\fontsize{7.2pt}{8.0pt}\selectfont \textbf{PSNR}$\uparrow$}
        & {\fontsize{7.2pt}{8.0pt}\selectfont \textbf{LPIPS}$\downarrow$}
        & {\fontsize{7.2pt}{8.0pt}\selectfont \textbf{Time}$\downarrow$}
        & {\fontsize{7.2pt}{8.0pt}\selectfont \textbf{SSIM}$\uparrow$}
        & {\fontsize{7.2pt}{8.0pt}\selectfont \textbf{PSNR}$\uparrow$}
        & {\fontsize{7.2pt}{8.0pt}\selectfont \textbf{LPIPS}$\downarrow$}
        & {\fontsize{7.2pt}{8.0pt}\selectfont \textbf{Time}$\downarrow$}
        & {\fontsize{7.2pt}{8.0pt}\selectfont \textbf{SSIM}$\uparrow$}
        & {\fontsize{7.2pt}{8.0pt}\selectfont \textbf{PSNR}$\uparrow$}
        & {\fontsize{7.2pt}{8.0pt}\selectfont \textbf{LPIPS}$\downarrow$}
        & {\fontsize{7.2pt}{8.0pt}\selectfont \textbf{Time}$\downarrow$}
        & {\fontsize{7.2pt}{8.0pt}\selectfont \textbf{SSIM}$\uparrow$}
        & {\fontsize{7.2pt}{8.0pt}\selectfont \textbf{PSNR}$\uparrow$}
        & {\fontsize{7.2pt}{8.0pt}\selectfont \textbf{LPIPS}$\downarrow$}
        & {\fontsize{7.2pt}{8.0pt}\selectfont \textbf{Time}$\downarrow$} \\
        \midrule
        COLMAP
        & 0.872 & 28.219 & 0.157 & 4.178\,m
        & 0.889 & 29.214 & 0.138 & 5.982\,m
        & 0.902 & 29.845 & 0.129 & 7.341\,m
        & 0.881 & 28.967 & 0.143 & 6.127\,m \\
        VGGT
        & 0.853 & 27.213 & 0.183 & 1.585\,s
        & 0.861 & 28.337 & 0.174 & 1.534\,s
        & 0.875 & 28.914 & 0.168 & 1.693\,s
        & 0.854 & 27.842 & 0.181 & 1.382\,s \\
        $\pi^3$
        & 0.874 & 28.179 & 0.153 & 1.464\,s
        & 0.884 & 29.102 & 0.142 & 1.487\,s
        & 0.895 & 29.613 & 0.135 & 1.446\,s
        & 0.876 & 28.721 & 0.152 & 1.764\,s \\
        \bottomrule
    \end{tabular}
    }}

    \vspace{0.42em}

    \makebox[\textwidth][c]{
    \resizebox{\textwidth}{!}{
    \begin{tabular}{l*{16}{c}}
        \toprule
        \multirow{2}{*}{\raisebox{-0.65ex}{\textbf{Method}}}
        & \multicolumn{4}{c}{\textbf{Tobacco\_Outdoor\_03}}
        & \multicolumn{4}{c}{\textbf{Tobacco\_Outdoor\_04}}
        & \multicolumn{4}{c}{\textbf{Tobacco\_Outdoor\_05}}
        & \multicolumn{4}{c}{\textbf{Tobacco\_Outdoor\_06}} \\
        \cmidrule(lr){2-5}
        \cmidrule(lr){6-9}
        \cmidrule(lr){10-13}
        \cmidrule(lr){14-17}
        & {\fontsize{7.2pt}{8.0pt}\selectfont \textbf{SSIM}$\uparrow$}
        & {\fontsize{7.2pt}{8.0pt}\selectfont \textbf{PSNR}$\uparrow$}
        & {\fontsize{7.2pt}{8.0pt}\selectfont \textbf{LPIPS}$\downarrow$}
        & {\fontsize{7.2pt}{8.0pt}\selectfont \textbf{Time}$\downarrow$}
        & {\fontsize{7.2pt}{8.0pt}\selectfont \textbf{SSIM}$\uparrow$}
        & {\fontsize{7.2pt}{8.0pt}\selectfont \textbf{PSNR}$\uparrow$}
        & {\fontsize{7.2pt}{8.0pt}\selectfont \textbf{LPIPS}$\downarrow$}
        & {\fontsize{7.2pt}{8.0pt}\selectfont \textbf{Time}$\downarrow$}
        & {\fontsize{7.2pt}{8.0pt}\selectfont \textbf{SSIM}$\uparrow$}
        & {\fontsize{7.2pt}{8.0pt}\selectfont \textbf{PSNR}$\uparrow$}
        & {\fontsize{7.2pt}{8.0pt}\selectfont \textbf{LPIPS}$\downarrow$}
        & {\fontsize{7.2pt}{8.0pt}\selectfont \textbf{Time}$\downarrow$}
        & {\fontsize{7.2pt}{8.0pt}\selectfont \textbf{SSIM}$\uparrow$}
        & {\fontsize{7.2pt}{8.0pt}\selectfont \textbf{PSNR}$\uparrow$}
        & {\fontsize{7.2pt}{8.0pt}\selectfont \textbf{LPIPS}$\downarrow$}
        & {\fontsize{7.2pt}{8.0pt}\selectfont \textbf{Time}$\downarrow$} \\
        \midrule
        COLMAP
        & 0.893 & 29.514 & 0.137 & 8.492\,m
        & 0.878 & 28.391 & 0.149 & 4.913\,m
        & 0.904 & 29.914 & 0.126 & 6.871\,m
        & 0.890 & 29.304 & 0.141 & 5.613\,m \\
        VGGT
        & 0.862 & 28.347 & 0.176 & 1.919\,s
        & 0.851 & 27.144 & 0.185 & 1.487\,s
        & 0.876 & 28.841 & 0.171 & 1.693\,s
        & 0.863 & 28.267 & 0.177 & 1.447\,s \\
        $\pi^3$
        & 0.887 & 29.289 & 0.141 & 1.631\,s
        & 0.873 & 28.256 & 0.155 & 1.603\,s
        & 0.899 & 29.712 & 0.132 & 1.592\,s
        & 0.886 & 29.187 & 0.148 & 1.729\,s \\
        \bottomrule
    \end{tabular}
    }}

    \vspace{0.42em}

    \makebox[\textwidth][c]{
    \resizebox{\textwidth}{!}{
    \begin{tabular}{l*{16}{c}}
        \toprule
        \multirow{2}{*}{\raisebox{-0.65ex}{\textbf{Method}}}
        & \multicolumn{4}{c}{\textbf{Wheat\_Indoor\_01}}
        & \multicolumn{4}{c}{\textbf{Wheat\_Field\_01}}
        & \multicolumn{4}{c}{\textbf{Wheat\_Field\_02}}
        & \multicolumn{4}{c}{\textbf{Soybean\_Indoor\_01}} \\
        \cmidrule(lr){2-5}
        \cmidrule(lr){6-9}
        \cmidrule(lr){10-13}
        \cmidrule(lr){14-17}
        & {\fontsize{7.2pt}{8.0pt}\selectfont \textbf{SSIM}$\uparrow$}
        & {\fontsize{7.2pt}{8.0pt}\selectfont \textbf{PSNR}$\uparrow$}
        & {\fontsize{7.2pt}{8.0pt}\selectfont \textbf{LPIPS}$\downarrow$}
        & {\fontsize{7.2pt}{8.0pt}\selectfont \textbf{Time}$\downarrow$}
        & {\fontsize{7.2pt}{8.0pt}\selectfont \textbf{SSIM}$\uparrow$}
        & {\fontsize{7.2pt}{8.0pt}\selectfont \textbf{PSNR}$\uparrow$}
        & {\fontsize{7.2pt}{8.0pt}\selectfont \textbf{LPIPS}$\downarrow$}
        & {\fontsize{7.2pt}{8.0pt}\selectfont \textbf{Time}$\downarrow$}
        & {\fontsize{7.2pt}{8.0pt}\selectfont \textbf{SSIM}$\uparrow$}
        & {\fontsize{7.2pt}{8.0pt}\selectfont \textbf{PSNR}$\uparrow$}
        & {\fontsize{7.2pt}{8.0pt}\selectfont \textbf{LPIPS}$\downarrow$}
        & {\fontsize{7.2pt}{8.0pt}\selectfont \textbf{Time}$\downarrow$}
        & {\fontsize{7.2pt}{8.0pt}\selectfont \textbf{SSIM}$\uparrow$}
        & {\fontsize{7.2pt}{8.0pt}\selectfont \textbf{PSNR}$\uparrow$}
        & {\fontsize{7.2pt}{8.0pt}\selectfont \textbf{LPIPS}$\downarrow$}
        & {\fontsize{7.2pt}{8.0pt}\selectfont \textbf{Time}$\downarrow$} \\
        \midrule
        COLMAP
        & 0.892 & 29.612 & 0.139 & 7.284\,m
        & 0.883 & 28.934 & 0.148 & 4.721\,m
        & 0.897 & 29.728 & 0.131 & 6.512\,m
        & 0.905 & 29.981 & 0.124 & 8.193\,m \\
        VGGT
        & 0.865 & 28.457 & 0.176 & 1.512\,s
        & 0.857 & 27.823 & 0.182 & 1.403\,s
        & 0.868 & 28.593 & 0.171 & 1.557\,s
        & 0.879 & 28.942 & 0.165 & 1.684\,s \\
        $\pi^3$
        & 0.889 & 29.471 & 0.147 & 1.683\,s
        & 0.879 & 28.713 & 0.155 & 1.516\,s
        & 0.892 & 29.501 & 0.138 & 1.746\,s
        & 0.901 & 29.718 & 0.129 & 1.533\,s \\
        \bottomrule
    \end{tabular}
    }}

    \vspace{0.42em}

    \makebox[\textwidth][c]{
    \resizebox{\textwidth}{!}{
    \begin{tabular}{l*{16}{c}}
        \toprule
        \multirow{2}{*}{\raisebox{-0.65ex}{\textbf{Method}}}
        & \multicolumn{4}{c}{\textbf{Soybean\_Indoor\_02}}
        & \multicolumn{4}{c}{\textbf{Wheat\_Outdoor\_02}}
        & \multicolumn{4}{c}{\textbf{Rapeseed\_Indoor\_01}}
        & \multicolumn{4}{c}{\textbf{Legume\_Indoor\_01}} \\
        \cmidrule(lr){2-5}
        \cmidrule(lr){6-9}
        \cmidrule(lr){10-13}
        \cmidrule(lr){14-17}
        & {\fontsize{7.2pt}{8.0pt}\selectfont \textbf{SSIM}$\uparrow$}
        & {\fontsize{7.2pt}{8.0pt}\selectfont \textbf{PSNR}$\uparrow$}
        & {\fontsize{7.2pt}{8.0pt}\selectfont \textbf{LPIPS}$\downarrow$}
        & {\fontsize{7.2pt}{8.0pt}\selectfont \textbf{Time}$\downarrow$}
        & {\fontsize{7.2pt}{8.0pt}\selectfont \textbf{SSIM}$\uparrow$}
        & {\fontsize{7.2pt}{8.0pt}\selectfont \textbf{PSNR}$\uparrow$}
        & {\fontsize{7.2pt}{8.0pt}\selectfont \textbf{LPIPS}$\downarrow$}
        & {\fontsize{7.2pt}{8.0pt}\selectfont \textbf{Time}$\downarrow$}
        & {\fontsize{7.2pt}{8.0pt}\selectfont \textbf{SSIM}$\uparrow$}
        & {\fontsize{7.2pt}{8.0pt}\selectfont \textbf{PSNR}$\uparrow$}
        & {\fontsize{7.2pt}{8.0pt}\selectfont \textbf{LPIPS}$\downarrow$}
        & {\fontsize{7.2pt}{8.0pt}\selectfont \textbf{Time}$\downarrow$}
        & {\fontsize{7.2pt}{8.0pt}\selectfont \textbf{SSIM}$\uparrow$}
        & {\fontsize{7.2pt}{8.0pt}\selectfont \textbf{PSNR}$\uparrow$}
        & {\fontsize{7.2pt}{8.0pt}\selectfont \textbf{LPIPS}$\downarrow$}
        & {\fontsize{7.2pt}{8.0pt}\selectfont \textbf{Time}$\downarrow$} \\
        \midrule
        COLMAP
        & 0.886 & 29.148 & 0.144 & 5.923\,m
        & 0.893 & 29.487 & 0.138 & 7.612\,m
        & 0.882 & 28.963 & 0.146 & 4.983\,m
        & 0.898 & 29.631 & 0.133 & 6.388\,m \\
        VGGT
        & 0.859 & 28.013 & 0.178 & 1.312\,s
        & 0.867 & 28.415 & 0.171 & 1.406\,s
        & 0.855 & 27.942 & 0.184 & 1.475\,s
        & 0.871 & 28.592 & 0.169 & 1.621\,s \\
        $\pi^3$
        & 0.881 & 28.934 & 0.151 & 1.728\,s
        & 0.889 & 29.315 & 0.145 & 1.557\,s
        & 0.879 & 28.798 & 0.152 & 1.629\,s
        & 0.894 & 29.508 & 0.140 & 1.507\,s \\
        \bottomrule
    \end{tabular}
    }}

    \vspace{0.42em}

    \makebox[\textwidth][c]{
    \resizebox{\textwidth}{!}{
    \begin{tabular}{l*{16}{c}}
        \toprule
        \multirow{2}{*}{\raisebox{-0.65ex}{\textbf{Method}}}
        & \multicolumn{4}{c}{\textbf{Pea\_Indoor\_01}}
        & \multicolumn{4}{c}{\textbf{Pea\_Indoor\_02}}
        & \multicolumn{4}{c}{\textbf{Pea\_Indoor\_03}}
        & \multicolumn{4}{c}{\textbf{Pea\_Indoor\_04}} \\
        \cmidrule(lr){2-5}
        \cmidrule(lr){6-9}
        \cmidrule(lr){10-13}
        \cmidrule(lr){14-17}
        & {\fontsize{7.2pt}{8.0pt}\selectfont \textbf{SSIM}$\uparrow$}
        & {\fontsize{7.2pt}{8.0pt}\selectfont \textbf{PSNR}$\uparrow$}
        & {\fontsize{7.2pt}{8.0pt}\selectfont \textbf{LPIPS}$\downarrow$}
        & {\fontsize{7.2pt}{8.0pt}\selectfont \textbf{Time}$\downarrow$}
        & {\fontsize{7.2pt}{8.0pt}\selectfont \textbf{SSIM}$\uparrow$}
        & {\fontsize{7.2pt}{8.0pt}\selectfont \textbf{PSNR}$\uparrow$}
        & {\fontsize{7.2pt}{8.0pt}\selectfont \textbf{LPIPS}$\downarrow$}
        & {\fontsize{7.2pt}{8.0pt}\selectfont \textbf{Time}$\downarrow$}
        & {\fontsize{7.2pt}{8.0pt}\selectfont \textbf{SSIM}$\uparrow$}
        & {\fontsize{7.2pt}{8.0pt}\selectfont \textbf{PSNR}$\uparrow$}
        & {\fontsize{7.2pt}{8.0pt}\selectfont \textbf{LPIPS}$\downarrow$}
        & {\fontsize{7.2pt}{8.0pt}\selectfont \textbf{Time}$\downarrow$}
        & {\fontsize{7.2pt}{8.0pt}\selectfont \textbf{SSIM}$\uparrow$}
        & {\fontsize{7.2pt}{8.0pt}\selectfont \textbf{PSNR}$\uparrow$}
        & {\fontsize{7.2pt}{8.0pt}\selectfont \textbf{LPIPS}$\downarrow$}
        & {\fontsize{7.2pt}{8.0pt}\selectfont \textbf{Time}$\downarrow$} \\
        \midrule
        COLMAP
        & 0.887 & 29.184 & 0.142 & 5.712\,m
        & 0.892 & 29.415 & 0.138 & 7.981\,m
        & 0.899 & 29.784 & 0.130 & 9.182\,m
        & 0.884 & 28.742 & 0.147 & 4.312\,m \\
        VGGT
        & 0.862 & 28.193 & 0.176 & 1.422\,s
        & 0.869 & 28.447 & 0.172 & 1.774\,s
        & 0.872 & 28.619 & 0.167 & 1.422\,s
        & 0.859 & 27.894 & 0.181 & 1.305\,s \\
        $\pi^3$
        & 0.884 & 29.041 & 0.148 & 1.638\,s
        & 0.889 & 29.294 & 0.144 & 1.582\,s
        & 0.895 & 29.593 & 0.137 & 1.693\,s
        & 0.881 & 28.593 & 0.153 & 1.445\,s \\
        \bottomrule
    \end{tabular}
    }}

    \vspace{0.42em}

    \makebox[\textwidth][c]{
    \resizebox{0.58\textwidth}{!}{
    \begin{tabular}{l*{8}{c}}
        \toprule
        \multirow{2}{*}{\raisebox{-0.65ex}{\textbf{Method}}}
        & \multicolumn{4}{c}{\textbf{Broccoli\_Indoor\_01}}
        & \multicolumn{4}{c}{\textbf{Wheat\_Indoor\_02}} \\
        \cmidrule(lr){2-5}
        \cmidrule(lr){6-9}
        & {\fontsize{7.2pt}{8.0pt}\selectfont \textbf{SSIM}$\uparrow$}
        & {\fontsize{7.2pt}{8.0pt}\selectfont \textbf{PSNR}$\uparrow$}
        & {\fontsize{7.2pt}{8.0pt}\selectfont \textbf{LPIPS}$\downarrow$}
        & {\fontsize{7.2pt}{8.0pt}\selectfont \textbf{Time}$\downarrow$}
        & {\fontsize{7.2pt}{8.0pt}\selectfont \textbf{SSIM}$\uparrow$}
        & {\fontsize{7.2pt}{8.0pt}\selectfont \textbf{PSNR}$\uparrow$}
        & {\fontsize{7.2pt}{8.0pt}\selectfont \textbf{LPIPS}$\downarrow$}
        & {\fontsize{7.2pt}{8.0pt}\selectfont \textbf{Time}$\downarrow$} \\
        \midrule
        COLMAP
        & 0.891 & 29.621 & 0.139 & 6.583\,m
        & 0.903 & 29.947 & 0.128 & 8.342\,m \\
        VGGT
        & 0.866 & 28.417 & 0.173 & 1.654\,s
        & 0.876 & 28.913 & 0.169 & 1.723\,s \\
        $\pi^3$
        & 0.887 & 29.478 & 0.146 & 1.578\,s
        & 0.900 & 29.815 & 0.135 & 1.568\,s \\
        \bottomrule
    \end{tabular}
    }}
\end{table*}

Front-end initialization represents the first bottleneck in the proposed phenotyping pipeline. Subsequent 3DGS densification, few-view enhancement, 2D-to-3D semantic back-projection, scale recovery, and leaf-level measurement all require stable camera parameters and initial geometry. To test whether 3DFMs can serve as the geometric entry point for plant 3D phenotyping, we compared COLMAP, VGGT, and $\pi^3$ on $26$ plants covering diverse crops, morphologies, and acquisition scenes. Table~\ref{tab:plant_26_compare} reports initialization efficiency and initialization-driven reconstruction quality using SSIM, PSNR, LPIPS, and front-end runtime. Runtime was restricted to initialization: feature extraction, matching, pose estimation, and sparse reconstruction for COLMAP; model inference plus bridge conversion to standardized camera parameters and sparse geometry for VGGT and $\pi^3$. We excluded downstream densification, scale recovery, and phenotypic measurement from the runtime evaluation.

The main gain was efficiency. COLMAP required an average front-end runtime of $6.52~\mathrm{min}$, with a median of $6.45~\mathrm{min}$. VGGT and $\pi^3$ required only $1.55~\mathrm{s}$ and $1.58~\mathrm{s}$ on average, with medians of $1.52~\mathrm{s}$ and $1.60~\mathrm{s}$, respectively. Based on mean runtime, VGGT and $\pi^3$ achieved approximately $252.6\times$ and $247.3\times$ speedups over COLMAP. Thus, 3DFM-based initialization compressed the dominant front-end cost from minutes to seconds, with consistent behavior across all samples.

This acceleration remained useful because the quality loss was limited, especially for $\pi^3$. COLMAP achieved the highest average SSIM, PSNR, and LPIPS values of $0.890$, $29.387~\mathrm{dB}$, and $0.138$. $\pi^3$ reached $0.886$, $29.191~\mathrm{dB}$, and $0.144$, remaining close to COLMAP. VGGT reached $0.863$, $28.333~\mathrm{dB}$, and $0.175$, showing a larger drop. Relative to COLMAP, $\pi^3$ reduced mean SSIM by only $0.004$, reduced PSNR by $0.196~\mathrm{dB}$, and increased LPIPS by $0.006$; VGGT reduced mean SSIM by $0.027$, reduced PSNR by $1.055~\mathrm{dB}$, and increased LPIPS by $0.036$. $\pi^3$ therefore provided an approximately $250\times$ acceleration with only a minor reconstruction-quality cost.

The comparison between $\pi^3$ and VGGT determined the default initializer for subsequent experiments. Both models enabled second-level feed-forward visual geometry inference, yet $\pi^3$ achieved higher SSIM and PSNR than VGGT on all $26$ samples, together with consistently lower LPIPS. This sample-wide advantage indicates more stable initialization for complex plant canopies, thin leaf boundaries, and local occlusions. Since subsequent 3DGS optimization, semantic back-projection, and leaf-level measurement depend on camera--point geometric consistency, $\pi^3$ provides a more suitable geometric starting point for the full pipeline.

Together, Table~\ref{tab:plant_26_compare} establishes 3DFM-based initialization as a practical alternative to COLMAP-style front ends in the proposed framework. COLMAP defines a high-quality conventional baseline, VGGT demonstrates the feasibility of unified feed-forward visual geometry inference for plant reconstruction, and $\pi^3$ provides the best quality--efficiency balance. With the front-end cost reduced from minutes to seconds, the next question is whether the 3DFM-driven pipeline can remain usable when the number of input views is further reduced.

\subsection{Comparison of performance in few-view reconstruction}

\begin{figure*}[!t]
    \centering
    \includegraphics[width=\linewidth]{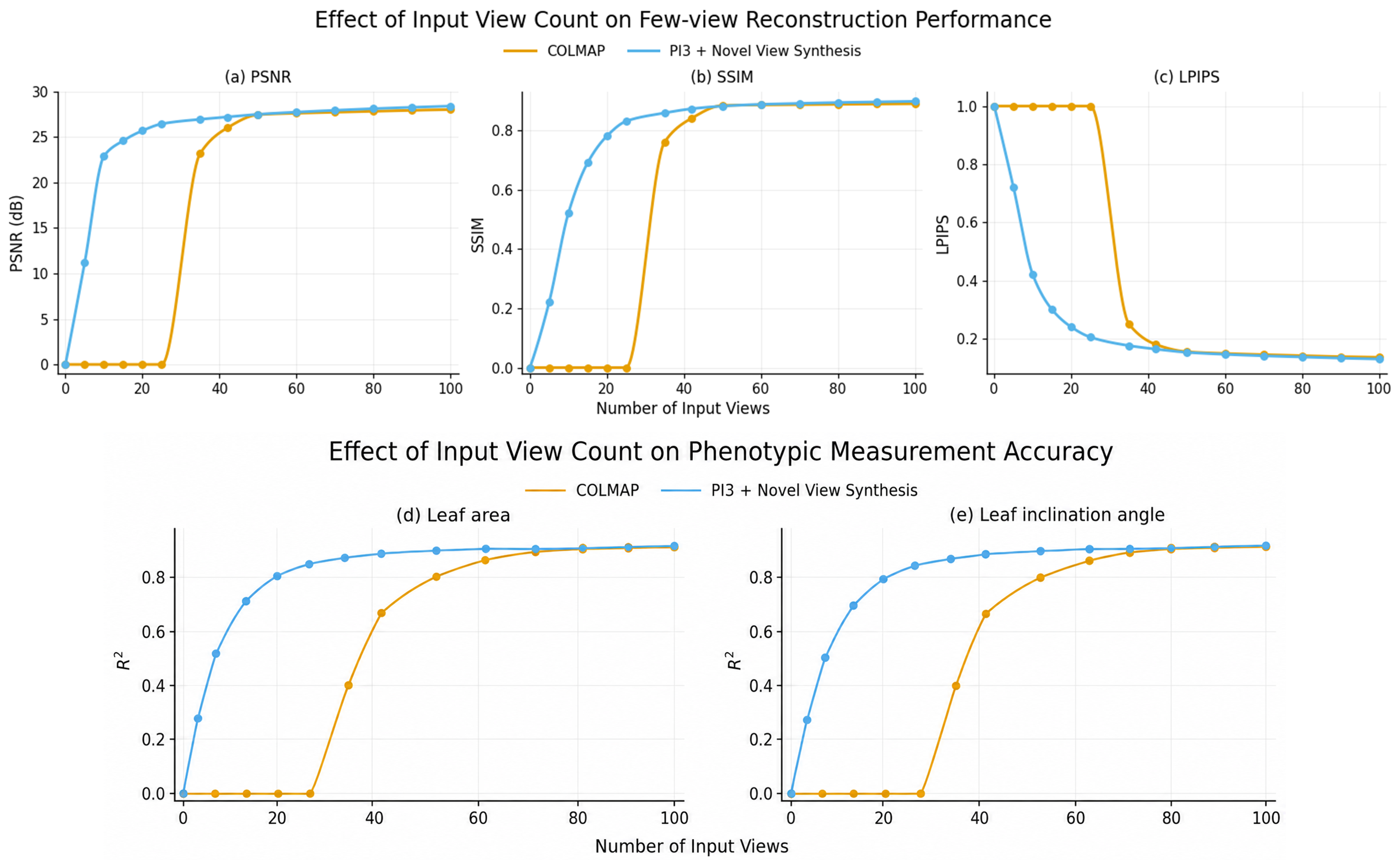}
    \caption{\textbf{Effect of Input View Count on Reconstruction and Phenotyping Performance.} PSNR, SSIM, and LPIPS are reported for COLMAP and the proposed $\pi^3$ + novel view synthesis pipeline under different numbers of input views. The proposed method achieves usable reconstruction at much lower view counts and approaches the dense-view performance ceiling more rapidly.}
    \label{fig:fewview}
\end{figure*}

After front-end initialization was compressed from the minute level to the second level, the bottleneck of high-throughput plant 3D phenotyping shifted further toward image acquisition. In field inspection, robotic platforms, and handheld rapid capture, view number and view overlap are often constrained by time, trajectory, and occlusion. To evaluate the usability of the proposed workflow under low-view input, we compared COLMAP versus the $\pi^3$ + novel-view synthesis pipeline across different input view counts, using reconstruction quality and phenotypic measurement accuracy as evaluation targets. The key question is whether stable 3D geometry can still be recovered under sparse views, then support leaf area and leaf inclination estimation.

As shown in Fig.~\ref{fig:fewview}, the reconstruction-quality curves show a clear view-threshold effect. In the low-view range, COLMAP failed to produce valid reconstructions, with PSNR and SSIM close to zero and LPIPS close to the failure ceiling, indicating that matching-based SfM cannot reliably establish camera poses and initial geometry under low-overlap input. The $\pi^3$ + novel-view synthesis pipeline recovered rapidly in the same range, with PSNR, SSIM, and LPIPS entering the usable range earlier. This difference indicates that the primary bottleneck in few-view reconstruction is whether initialization can be established; $\pi^3$ provides a stable front-end geometry, while novel-view synthesis and refinement supplement missing observations, thereby reducing the number of views required for usable reconstruction.

As the view count increased, COLMAP showed a sharp improvement at approximately $30$--$45$ views, indicating that conventional SfM requires sufficient view overlap to enter a stable reconstruction regime. The $\pi^3$ + novel-view synthesis pipeline had already approached a plateau before this threshold. After more than $50$ views, the PSNR, SSIM, and LPIPS curves of the two methods gradually converged. This trend shows that the main benefit of the proposed method lies in the low-view regime: its core value is shifting the usable reconstruction threshold forward, rather than raising the performance ceiling under dense-view input.

The phenotypic measurement curves further show that the few-view advantage propagates to terminal traits. The $R^2$ values of leaf area and leaf inclination angle increased with view count. COLMAP could not produce valid phenotypic estimates in the reconstruction-failure range, whereas the $\pi^3$ + novel-view synthesis pipeline reached higher accuracy with fewer views. The increase in leaf-area $R^2$ reflects more stable scale recovery, leaf boundaries, and instance separation. The increase in inclination-angle $R^2$ indicates more reliable leaf-surface geometry, vertical reference, and normal estimation. Overall, this experiment shows that the proposed workflow reduces the view threshold required for both 3D reconstruction and organ-level phenotypic measurement, making it suitable for low-cost rapid acquisition.

\subsection{2D-to-3D semantic transfer achieves more consistent plant organ segmentation than direct 3D approaches}

Following reconstruction enhancement and point cloud densification, we compared two segmentation paradigms for reconstructed plant point clouds: direct 3D semantic segmentation and projection-based 2D-to-3D semantic transfer. The 3D baselines included PSegNet, TPointNet++, and PointTransformerV3, whereas the 2D-to-3D branch used the proposed pipeline of multi-view rendering, label back-projection, and multi-view fusion. For training the 3D baselines, we constructed a cross-crop dataset by integrating SoybeanMVS~\citep{sun2023soybeanmvs}, MaizeField3D~\citep{kimara2026maizefield3d}, and syau-single-maize~\citep{yang2024maize}, and split it into training and validation sets at an 8:2 ratio to support multi-crop segmentation with a single model. We then evaluated all trained models on our reconstructed plant dataset.

\begin{figure*}[!t]
    \centering
    \includegraphics[width=1.0\linewidth]{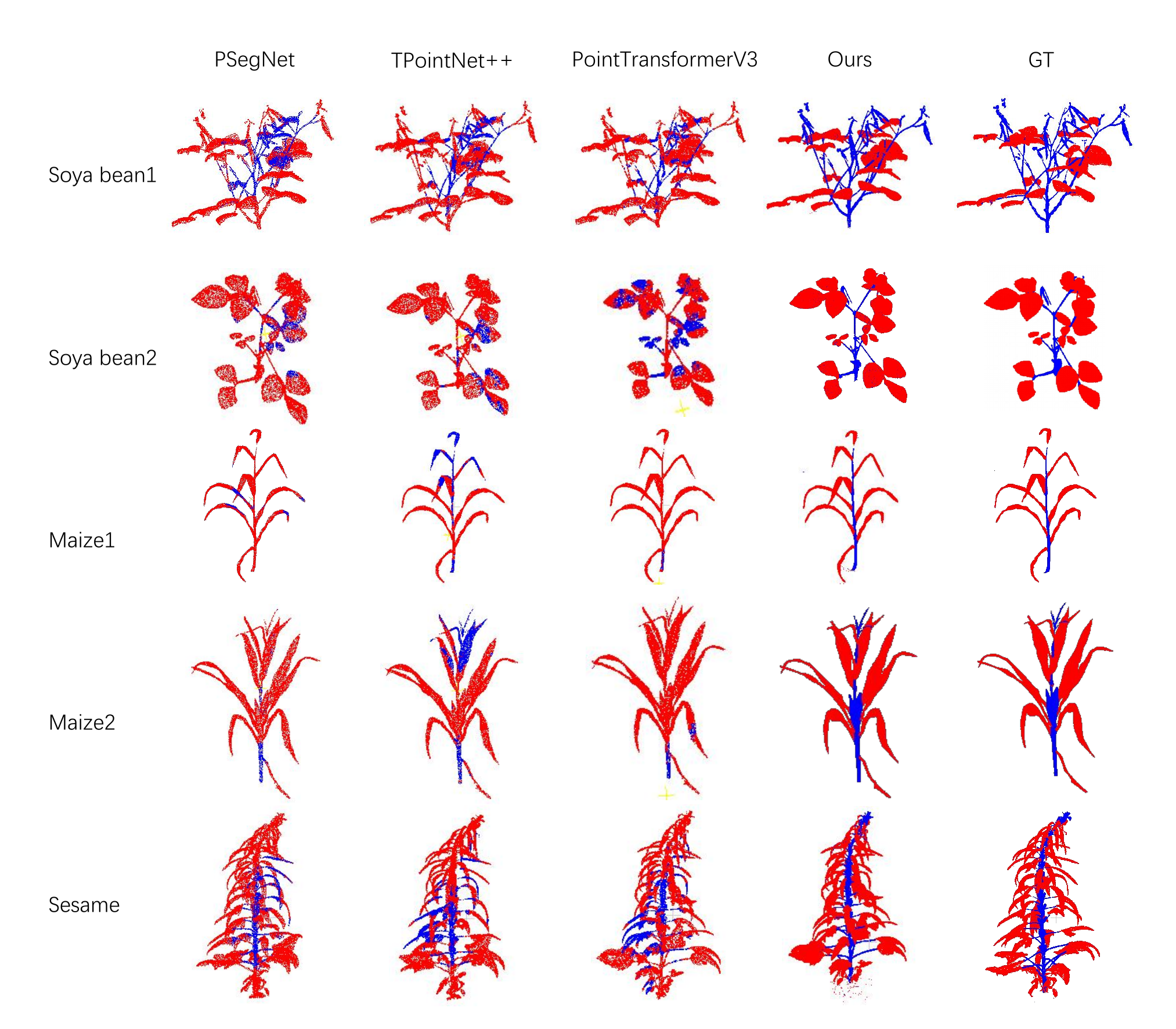}
    \caption{\textbf{Qualitative comparison of plant organ segmentation results across methods.}
    Qualitative results of PSegNet, TPointNet++, PointTransformerV3, and the proposed method on soybean, maize, and sesame samples, with GT shown for reference.}
    \label{fig:segmentation_qualitative}
\end{figure*}
\FloatBarrier

As shown in Fig.~\ref{fig:segmentation_qualitative}, Direct 3D segmentation showed clear limitations when applied to plant point cloud scenarios. Boundaries between leaves and stems were frequently blurred by thin organs, severe occlusion, uneven point density, and local holes, resulting in fragmentation, incorrect merging, and class confusion. Cross-crop structural variation further reduced stability, since maize exhibits a stronger vertical organization, whereas soybean and sesame contain denser branching and more entangled organs.

The projection-based 2D-to-3D strategy produced more consistent results across crops. The densified point cloud was first rendered into multiple views, and a pixel-to-point index map was constructed to establish explicit correspondence between image space and 3D space. Semantic predictions were then obtained in 2D, back-projected to the point cloud, and fused across views. This design preserved leaf boundaries more effectively, improved robustness under occlusion, and reduced the ambiguity introduced by post-hoc nearest-neighbor reprojection.

\begin{table*}[!t]
\centering
\caption{\textbf{Quantitative comparison of plant organ segmentation performance on different crops.}}
\label{tab:segmentation_quantitative}

\footnotesize
\renewcommand{\arraystretch}{1.18}
\setlength{\tabcolsep}{8pt}

\resizebox{0.95\textwidth}{!}{
\begin{tabular}{llcccc}
\toprule
\textbf{Crop} &
\textbf{Method} &
\textbf{OA$\uparrow$} &
\textbf{mIoU$\uparrow$} &
\textbf{Leaf IoU$\uparrow$} &
\textbf{Stem IoU$\uparrow$} \\
\midrule

\multirow{4}{*}{Soybean1}
& PSegNet            & 0.7028 & 0.4656 & 0.6673 & 0.2639 \\
& TPointNet++        & 0.7561 & 0.5144 & 0.7275 & 0.3014 \\
& PointTransformerV3 & 0.7676 & 0.5174 & 0.7429 & 0.2920 \\
& \textbf{Ours}      & \textbf{0.9660} & \textbf{0.9185} & \textbf{0.9543} & \textbf{0.8827} \\
\midrule

\multirow{4}{*}{Soybean2}
& PSegNet            & 0.7524 & 0.4164 & 0.7465 & 0.0863 \\
& TPointNet++        & 0.7364 & 0.3914 & 0.7327 & 0.0501 \\
& PointTransformerV3 & 0.6337 & 0.3504 & 0.6219 & 0.0790 \\
& \textbf{Ours}      & \textbf{0.9362} & \textbf{0.8216} & \textbf{0.9237} & \textbf{0.7195} \\
\midrule

\multirow{4}{*}{Maize1}
& PSegNet            & 0.7809 & 0.3962 & 0.7803 & 0.0121 \\
& TPointNet++        & 0.7880 & 0.4891 & 0.7759 & 0.2023 \\
& PointTransformerV3 & 0.9075 & 0.5073 & 0.9065 & 0.1081 \\
& \textbf{Ours}      & \textbf{0.9962} & \textbf{0.9810} & \textbf{0.9957} & \textbf{0.9662} \\
\midrule

\multirow{4}{*}{Maize2}
& PSegNet            & 0.9222 & 0.5943 & 0.9199 & 0.2688 \\
& TPointNet++        & 0.7821 & 0.4411 & 0.7763 & 0.1060 \\
& PointTransformerV3 & 0.9062 & 0.5162 & 0.9049 & 0.1275 \\
& \textbf{Ours}      & \textbf{0.9956} & \textbf{0.9816} & \textbf{0.9950} & \textbf{0.9682} \\
\midrule

\multirow{4}{*}{Sesame}
& PSegNet            & 0.8482 & 0.5899 & 0.8351 & 0.3446 \\
& TPointNet++        & 0.7847 & 0.4999 & 0.7699 & 0.2299 \\
& PointTransformerV3 & 0.7664 & 0.4854 & 0.7498 & 0.2209 \\
& \textbf{Ours}      & \textbf{0.9742} & \textbf{0.9096} & \textbf{0.9698} & \textbf{0.8495} \\
\bottomrule

\end{tabular}
}
\end{table*}

The quantitative comparison in Table~\ref{tab:segmentation_quantitative} further supports these observations. The proposed method consistently achieved the best OA, mIoU, Leaf IoU, and Stem IoU across all evaluated crops. Overall, the advantage of 2D-to-3D segmentation lies in its better match to reconstructed plant data. It reduces dependence on large-scale point-level annotations, transfers more reliably across crops, and handles thin, overlapping structures more effectively through multi-view fusion and explicit geometric correspondence. We therefore adopt 2D-to-3D semantic transfer as the core strategy for subsequent leaf instance separation and phenotypic analysis.

\subsection{Leaf area and inclination angle are reliably estimated from reconstructed 3D plant structures}

After reconstruction and segmentation, the final test of the pipeline is whether reconstructed and segmented 3D structures can be converted into reliable organ-level traits. We evaluated leaf area and leaf inclination angle as terminal indicators. Leaf area depends on metric scale, leaf boundary preservation, instance separation, and mesh-based area computation. Leaf inclination angle depends on leaf-surface continuity, the vertical reference, and single-leaf normal estimation. These two traits evaluate scale-related and posture-related phenotypes, respectively.

\begin{figure*}[!t]

    \centering
    \includegraphics[width=\linewidth]{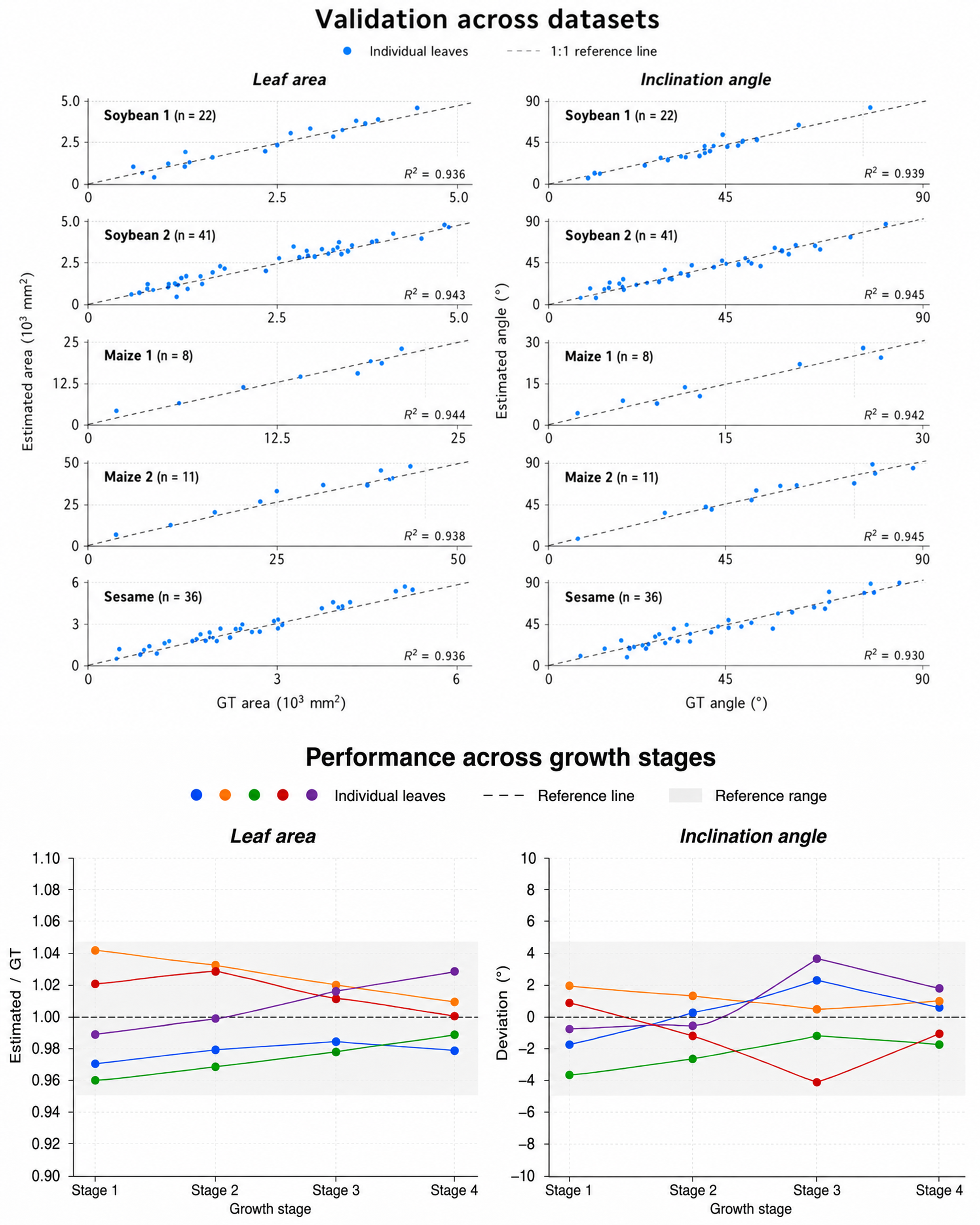}
    \caption{\textbf{Phenotypic measurement accuracy across crops and growth stages.}
    Top: cross-crop agreement between estimated and manual leaf area and inclination angle. Bottom: stage-wise leaf-area Estimated/GT ratios and signed leaf-inclination deviations, with shaded reference ranges shown for area ratios and inclination deviations, respectively.}
    \label{fig:phenotype_accuracy}
\end{figure*}

Across five representative soybean, maize, and sesame plants, the estimated traits remained highly consistent with manual measurements (Fig.~\ref{fig:phenotype_accuracy}). Leaf-area regression reached $R^2=0.9362$--$0.9438$, and leaf-inclination regression reached $R^2=0.9307$--$0.9455$, indicating that the pipeline preserved inter-leaf differences in both size and orientation. At the plant level, the $\mathrm{Estimated}/\mathrm{GT}$ ratio of total leaf area ranged from $0.9514$ to $1.0629$, corresponding to an approximately $-4.86\%$ to $+6.29\%$ deviation. The ratios remained close to $1.0$, suggesting limited global scale distortion. The mean absolute error of leaf inclination angle was approximately $2.04^\circ$, indicating stable coordinate normalization and normal estimation for posture traits.

Crop-specific patterns reflected structural differences in measurement difficulty. Maize leaves are elongated and axis-dominant, favoring boundary preservation and normal estimation; maize1 reached $R^2=0.9438$ for leaf area, and maize2 reached $R^2=0.9449$ for inclination angle. Soybean showed more overlap, petiole connection, and local occlusion; Soybean1 had the lowest leaf-area regression, with $R^2=0.9362$, reflecting greater boundary-recovery difficulty in compound leaves. Sesame maintained $R^2=0.9363$ for leaf area and $R^2=0.9307$ for inclination angle, supporting transfer across different leaf forms and canopy structures.

\FloatBarrier

Stage-wise results further showed measurement stability. Across four growth stages, the $\mathrm{Estimated}/\mathrm{GT}$ ratio for leaf area fluctuated around $1.0$, indicating no cumulative area error with plant growth and leaf expansion. Inclination-angle deviations mostly stayed within the $\pm 5^\circ$ reference range, indicating stable vertical-reference recovery and leaf-normal estimation across stages. This result extends the validation from single-time-point cross-crop measurement to continuous growth-stage phenotyping, suggesting that the proposed pipeline has the potential to support dynamic phenotypic recording. Together, the cross-crop regression, stage-wise stability, and low angular error show that the pipeline preserves the geometric, semantic, and metric consistency required for organ-level measurement, enabling low-cost image-based reconstruction to be converted into comparable 3D plant phenotypic traits.

\subsection{All three design choices improve performance, with sparse initialization contributing the strongest gain}

Table~\ref{tab:ablation} presents ablation results showing consistent gains from all three design choices, with the strongest effect arising from sparse initialization. Under matched dense modeling and augmentation settings, $\pi^3$ improves PSNR, area $R^2$, and angle $R^2$ over COL while also reducing the overall runtime. Mip consistently outperforms standard 3DGS in paired configurations, indicating better preservation of thin leaf boundaries and local structural continuity. Novel-view augmentation brings smaller but systematic gains, mainly by compensating for sparse observations and refining local geometry.

\newcommand{\headcell}[1]{
  \begin{tabular}[c]{@{}c@{}}\bfseries\boldmath #1\end{tabular}
}

\begin{table*}[!t]
\centering
\caption{\textbf{Ablation study of the proposed framework with module-wise runtime.}}
\label{tab:ablation}

\footnotesize
\renewcommand{\arraystretch}{1.18}
\setlength{\tabcolsep}{4pt}

\resizebox{\textwidth}{!}{
\begin{tabular}{ccccccccccc}
\toprule
\textbf{No.} &
\makecell{\textbf{Sparse}\\\textbf{init.}} &
\makecell{\textbf{Dense}\\\textbf{model}} &
\makecell{\textbf{Novel-view}\\\textbf{aug.}} &
\textbf{PSNR$\uparrow$} &
\makecell{\textbf{Area}\\\textbf{$R^2\uparrow$}} &
\makecell{\textbf{Angle}\\\textbf{$R^2\uparrow$}} &
\makecell{\textbf{Init.}\\\textbf{time}} &
\makecell{\textbf{Dense}\\\textbf{time}\\\textbf{(min)}} &
\makecell{\textbf{Aug.}\\\textbf{time}\\\textbf{(min)}} &
\makecell{\textbf{Total}\\\textbf{time}\\\textbf{(min)}$\downarrow$} \\
\midrule
1 & COL     & 3DGS & $\times$      & 27.98 & 0.898 & 0.910 & 6.4 min & 30.9 & 0.0 & 37.3 \\
2 & COL     & 3DGS & $\checkmark$  & 28.24 & 0.905 & 0.917 & 6.4 min & 30.9 & 1.2 & 38.5 \\
3 & COL     & Mip  & $\times$      & 28.56 & 0.916 & 0.925 & 6.4 min & 31.5 & 0.0 & 37.9 \\
4 & COL     & Mip  & $\checkmark$  & 28.95 & 0.924 & 0.934 & 6.4 min & 31.5 & 1.2 & 39.1 \\
5 & $\pi^3$ & 3DGS & $\times$      & 29.04 & 0.919 & 0.926 & 1.6 s   & 30.8 & 0.0 & \textbf{30.8} \\
6 & $\pi^3$ & 3DGS & $\checkmark$  & 29.30 & 0.926 & 0.932 & 1.6 s   & 30.8 & 1.2 & 32.0 \\
7 & $\pi^3$ & Mip  & $\times$      & 29.58 & 0.935 & 0.940 & 1.6 s   & 31.4 & 0.0 & 31.4 \\
8 & $\pi^3$ & Mip  & $\checkmark$  & \textbf{29.86} & \textbf{0.941} & \textbf{0.946} & 1.6 s & 31.4 & 1.2 & 32.6 \\
\bottomrule
\end{tabular}
}

\vspace{2pt}
\begin{minipage}{0.98\textwidth}
\footnotesize
\textit{Note:} COL denotes COLMAP initialization. $\pi^3$ initialization is reported in seconds, whereas dense reconstruction, novel-view augmentation, and total runtime are reported in minutes. Dense time is computed as total time minus sparse initialization and novel-view augmentation time, and is rounded to one decimal place.
\end{minipage}
\end{table*}

At the configuration level, No.8 ($\pi^3$ + Mip + novel-view augmentation) achieves the best PSNR, area $R^2$, and angle $R^2$, demonstrating that the framework's advantage stems from the coupling of stable initialization, measurement-oriented densification, and sparse-view enhancement rather than any single module alone. Notably, No.5 gives the shortest runtime, whereas No.8 gives the highest accuracy, indicating that the added cost of novel-view augmentation is limited but sufficient to further release the potential of the $\pi^3$--Mip combination.

\section{Discussion}

\subsection{From 3DFM-based initialization to measurable phenotyping}

This study positions visual geometry 3DFMs as the front-end geometric foundation for plant 3D phenotyping and tests whether this change propagates to organ-level measurement. COLMAP-style initialization relies on feature extraction, matching, and incremental pose recovery, making it vulnerable to repetitive texture, occlusion, and thin organs. In contrast, $\pi^3$ reduces the average front-end runtime from $6.52~\mathrm{min}$ to $1.58~\mathrm{s}$ while maintaining reconstruction quality close to COLMAP, providing an efficient geometric entry point for 3DGS densification, few-view enhancement, and 3D semantic perception.

This front-end shift also relaxes acquisition constraints. Combined with novel-view synthesis, $\pi^3$ lowers the usable reconstruction threshold and recovers stable geometry from fewer views. The pipeline links faster initialization, synthesized observations, boundary-preserving 3DGS, 2D-to-3D semantic transfer, scale recovery, and leaf instance separation, converting reconstructed geometry into metric leaf-level objects. Thus, measurable traits require not only visual fidelity, but also geometric continuity, semantic consistency, and physical scale.

The phenotypic results verify this system-level propagation. Across representative crops, the plant-level $\mathrm{Estimated}\text{-}\mathrm{to}\text{-}\mathrm{GT}$ ratio of total leaf area ranges from $0.9514$ to $1.0629$, and the mean absolute error of leaf inclination angle is approximately $2.04^\circ$. The near-unity area ratios and low angular error indicate limited scale bias, stable vertical reference recovery, leaf-surface reconstruction, and normal estimation. Thus, 3DFM-based initialization establishes a verifiable geometric starting point from low-cost images to reconstruction, perception, and organ-level phenotypic extraction.

\subsection{Current limitations and applicability boundaries}

The current framework is most reliable for single-plant, close-range, closed-loop acquisition or structurally separated scenes, where viewpoint continuity and target visibility support coherent geometry across initialization, densification, semantic transfer, and measurement. Dense canopies, dynamic disturbances, cluttered backgrounds, and persistent occlusion remain challenging for camera recovery, local completion, cross-view semantic consistency, and leaf instance separation. Therefore, the present results mainly support low-cost, close-range, isolated-plant phenotyping, with further validation required in field-scale canopies.

Metric scale recovery is another boundary. Absolute traits such as leaf area and plant height require a reliable scale anchor. This study uses pot geometry as an in-scene reference, reducing calibration cost in potted or controlled acquisition, but this is only one anchoring strategy. In field deployment, scale may come from platform pose, camera height, ground-plane constraints, row or plant spacing, or reference markers. Without such an anchor, the workflow supports relative structural comparison and temporal change analysis rather than absolute traits.

Local organ separation remains challenging. Leaf-area and leaf-inclination errors are affected by edge completeness, petiole connections, leaf contact, occlusion, and curled structures, which complicate 2D-to-3D label transfer, instance separation, and mesh reconstruction. Although current results show stable overall accuracy, complex canopies still require leaf-level error analysis, failure visualization, and larger-scale validation. Future work should address marker-free scale recovery, robust organ separation, and higher-order phenotypes such as curvature, torsion, stem topology, and temporal growth dynamics. At this stage, the framework should be regarded as validated for low-cost, close-range, organ-level 3D phenotyping, with dense field canopies and marker-free deployment requiring further validation.

\subsection{Implications for low-cost and cross-crop plant phenotyping}

Within the applicability boundaries discussed above, this study shows that smartphone images can support measurable organ-level 3D plant phenotyping. Low-cost acquisition, 3DFM-based front-end initialization, and few-view reconstruction reduce hardware, time, and data barriers, while geometry-constrained 3DGS, 2D-to-3D semantic transfer, scale recovery, and leaf instance separation convert image-based reconstruction into measurement-ready objects with physical scale and organ boundaries. Thus, low-cost 3D phenotyping depends on coordinated reductions in acquisition, reconstruction, and measurement conversion.

To our knowledge, this study is among the first to systematically introduce visual geometry 3DFMs into plant 3D phenotyping and validate them as a front-end geometric foundation linking image acquisition, 3D reconstruction, and organ-level measurement. In this framework, 3DFMs provide a unified entry point for camera recovery, initial structure formation, few-view completion, and downstream 3D optimization. Accordingly, plant 3D phenotyping evaluation should extend beyond rendering quality to semantic consistency, metric scale, organ-instance quality, and final trait accuracy.

The cross-crop dataset and manually annotated ground truth support system-level evaluation beyond reconstruction quality, covering few-view stability, 3D perception, and organ-level measurement. Overall, this study establishes a 3DFM-driven, measurement-oriented route for low-cost cross-crop 3D plant phenotyping, connecting second-level initialization, few-view reconstruction, semantic perception, dataset-supported validation, and organ-level trait extraction into a complete evidence chain. The framework remains mainly suited to close-range, single-plant, or structurally separated acquisition, and future extensions are needed for complex field canopies, marker-free scale recovery, and higher-order dynamic traits.

\section{Conclusions}

This study presents a 3DFM-driven workflow for low-cost, cross-crop 3D plant phenotyping. By replacing COLMAP-style matching-based initialization with feed-forward 3D foundation model inference, the proposed pipeline compresses the reconstruction front end from minutes to seconds while maintaining reconstruction quality close to conventional methods. Combined with Mip-Splatting-based densification, few-view supplementation, 2D-to-3D semantic transfer, metric scale recovery, and leaf instance separation, the framework converts low-cost image inputs into measurable organ-level 3D traits. Experiments across diverse crops and acquisition conditions demonstrate that the proposed route supports rapid reconstruction, robust segmentation, and reliable estimation of leaf area and inclination angle. Future work should further extend the method to dense field canopies, marker-free scale recovery, stronger organ separation under occlusion, and multi-temporal dynamic phenotyping.

\section*{CRediT authorship contribution statement}
Wei Zhou and Hanyue Jia contributed equally to this work. All authors jointly conceived and designed the study. Hanyue Jia and Wei Zhou developed the methodology and performed the experiments. Wenbo Zhou and Yanan Li contributed to data analysis and experimental validation. Hanyue Jia and Wei Zhou prepared the initial manuscript. Hao Lu and Tingting Wu supervised the study, reviewed the manuscript, and provided critical revisions. All authors read and approved the final manuscript.

\section*{Declaration of competing interest}
The authors declare that they have no known competing financial interests or personal relationships that could have appeared to influence the work reported in this paper.

\section*{Funding}
This work was supported by the National Natural Science Foundation of China under Grant No. 62576146.

\section*{Acknowledgements}
The authors gratefully acknowledge the financial support from the National Natural Science Foundation of China.



\bibliographystyle{elsarticle-harv}
\bibliography{refs}

\end{document}